\documentclass[sigconf, natbib=true, anonymous=false]{acmart}
\usepackage{dsfont}
\AtBeginDocument{%
  }

\setcopyright{acmlicensed}
\copyrightyear{2018}
\acmYear{2018}
\acmDOI{XXXXXXX.XXXXXXX}

\acmConference[SIGIR'25]{July 13--18, 2025}{Padua, Italy}

\acmISBN{978-1-4503-XXXX-X/18/06}

\usepackage{multirow}
\usepackage{setspace}
\usepackage{algorithm}
\usepackage{algorithmic}
\usepackage[algo2e, ruled,linesnumbered]{algorithm2e} 
\usepackage{amsmath}
\usepackage{xcolor}
\usepackage[section]{placeins}
\usepackage{titlesec}  

\titlespacing{\section}{0pt}{0.5ex}{0pt}  
\titlespacing{\subsection}{0pt}{0.5ex}{0pt}
\begin{document}


\title{Homophily-aware Heterogeneous Graph Contrastive Learning}

\author{Haosen Wang}
\affiliation{%
  \institution{Southeast University;}
  \country{China}
}

\author{Chenglong Shi}
\affiliation{%
  \institution{Southeast University}
  \city{Nanjing}
  \country{China}
}

\author{Can Xu}
\affiliation{%
  \institution{East China Normal University}
  \city{Shanghai}
  \country{China}}

\author{Surong Yan}
\affiliation{%
  \institution{Zhejiang University of Finance and Economics}
  \city{Hangzhou}
  \country{China}}

\author{Pan Tang}
\authornotemark[1]
\affiliation{%
  \institution{Southeast University}
  \city{Nanjing}
  \country{China}}

\renewcommand{\abstractname}{ABSTRACT}
\begin{abstract}
Heterogeneous graph pre-training (HGP) has demonstrated remarkable performance across various domains. However, the issue of heterophily in real-world heterogeneous graphs (HGs) has been largely overlooked. To bridge this research gap, we proposed a novel heterogeneous graph contrastive learning framework, termed HGMS, which leverages connection strength and multi-view self-expression to learn homophilous node representations. Specifically, we design a heterogeneous edge dropping augmentation strategy that enhances the homophily of augmented views. Moreover, we introduce a multi-view self-expressive learning method to infer the homophily between nodes. In practice, we develop two approaches to solve the self-expressive matrix. The solved self-expressive matrix serves as an additional augmented view to provide homophilous information and is used to identify false negatives in contrastive loss. Extensive experimental results demonstrate the superiority of HGMS across different downstream tasks. 

\end{abstract}

\patchcmd{\maketitle}{\@copyrightspace}{\@copyrightspace\vspace{-3ex}\noindent\textbf{CCS CONCEPTS}}{}{}

\ccsdesc[500]{Computing methodologies~Machine learning}

\renewcommand{\keywordsname}{KEYWORDS}

\keywords{Heterogeneous graph, Contrastive learning, Self-expressive learning}

\received{20 February 2007}
\received[revised]{12 March 2009}
\received[accepted]{5 June 2009}

\maketitle

\section{INTRODUCTION}
Heterogeneous graphs (HGs) are prevalent in real-world scenarios, providing a powerful tool to model diverse objects and complex semantic relationships, such as academic networks \cite{HAN, HeCo}, and fraud networks \cite{fraud1, fraud2}, recommendation networks \cite{rec1, rec2} and so on. Heterogeneous graph neural networks (HGNNs) have attracted significant attention for their capacity to address the heterogeneity in graphs \cite{MAGNN, drug1}. 

Recently, heterogeneous graph pre-training (HGP) using self-supervised learning (SSL) techniques, such as contrastive learning, has emerged as a promising research direction. This solution reduces reliance on labeled data by extracting supervised signals directly from raw data. Among HGP methods, those based on metapaths have become a mainstream approach \cite{HDMI, DMGI, HeCo, HGCML}. These methods utilize predefined metapaths to construct homogeneous subgraphs centered around specific node types, facilitating the extraction of rich semantic relationships. 
Actually, metapaths are used to capture the homophily (i.e., interactions and connections amongst nodes of the same category) in HGs. For example, in an academic network, two papers co-authored by the same author are likely to belong to the same research field, corresponding to the metapath "paper-author-paper". However, many real-world HGs exhibit low homophily (i.e., heterophily), where nodes connected by metapaths may belong to different categories. Take the movie network as an example: an actor can appear in various genres, and a screenwriter can craft scripts across different genres. 

In this study, we conduct an empirical analysis (shown in Section \ref{section: Empirical study 1}) to explore the impact of homophily on HGP methods. Our findings reveal that the performance of HGP is closely associated with the homophily of the metapath-based subgraphs. Specifically, HGP methods generally perform better on HGs with high homophily in metapath-based subgraphs, whereas their performance deteriorates when the subgraphs exhibit low homophily. This highlights the importance of enhancing the homophily of metapath-based subgraphs and mitigating the influence of heterophilic structures to improve the expressive power of HGP models.

To address the challenge of heterophily in HGs, we propose a novel heterogeneous graph contrastive learning framework, called HGMS, which leverages connection strength and multi-view self-expression. Through empirical study (shown in Section \ref{section: Empirical study 2}), we observe that metapath-based neighbor pairs with higher connection strength tend to exhibit higher homophily. To exploit this insight, we design a heterogeneous edge dropping augmentation strategy that enhances the homophily of augmented views. This strategy preserves metapath-based connections with stronger connection strength whenever possible, and we provide both mathematical and experimental evidence to demonstrate its effectiveness. Furthermore, we introduce a multi-view self-expressive learning method to infer the homophily between nodes. In practice, we propose two approaches for solving the self-expressive matrix: the \textit{\textbf{multi-view self-expressive network}} and the \textit{\textbf{closed-form solution}}. The learned self-expressive matrix plays a dual role in enhancing the homophily of node representations. First, it is sparsified and treated as an augmented view, guiding the representation learning process. Second, it acts as a measure of the likelihood that negative sample pairs are false, i.e., they are actually positive pairs. Our theoretical analysis shows that the tailored contrastive loss, which mitigates false negatives through self-expression, provides a tighter lower bound on the mutual information (MI) between raw node features and embeddings from augmented views.

The main contributions of this study are summarized as follows:

(1) We conduct an empirical study demonstrating that homophily significantly impacts the performance of HGP methods.

(2) We design a heterogeneous edge-dropping strategy for HG augmentation, which enhances the homophily of augmented views.

(3) We propose a multi-view self-expressive learning method to capture the homophily among nodes, with the learned self-expressive matrix serving dual roles: providing additional augmented views and mitigating the effects of false negatives.

(4) Extensive experiments on six public datasets validate the effectiveness of the proposed models, showing superior performance in various downstream tasks.

\section{RELATED WORK}
\subsection{Heterogeneous Graph Neural Networks}
Heterogeneous Graph Neural Networks (HGNNs) \cite{HeGNN, HAN, Hetero} have gained extensive attention, with current research categorized into metapath-based and metapath-free methods. Metapath-based methods leverage metapaths to extract high-order semantics from HG. For instance, HAN \cite{HAN} introduces node-level and semantic-level attention mechanisms to capture hierarchical features. MAGNN \cite{MAGNN} incorporates all nodes in metapath instances to mine more intrinsic information. SeHGNN \cite{SeHGNN} neighbor aggregation during the preprocessing stage using a mean aggregator, significantly improving training efficiency. HDHGR \cite{HDHGR} explores heterophily in HGs for the first time, calculating attribute and label similarity to rewire metapath subgraphs. Hetero$^{2}$Net \cite{Hetero} introduce both masked metapath prediction and masked label prediction tasks to handle address homophilic HGs. In contrast, metapath-free methods directly aggregate features from different node types.  HGT \cite{HGT} is a typical example, designing a heterogeneous mutual attention mechanism to differentiate messages from various meta-relations. Simple-HGN \cite{Simple-HGN} enhances GAT for HGs by introducing learnable edge-type embeddings, residual connections, and L2 normalization. Both HINormer \cite{HINormer} and PHGT \cite{PHGT} are heterogeneous graph transformer frameworks designed to learn node representations. 


\subsection{Heterogeneous Graph Pre-training}
Given the high cost of data annotation, heterogeneous graph pre-training (HGP) has gradually emerged as an effective solution. Existing HGP models can be broadly categorized into contrastive and generative methods. Contrastive HGP models capture node characteristics by maximizing/minimizing the consistency between these positive/negative pairs. HDGI \cite{HDGI},  DMGI \cite{DMGI}, and HDMI \cite{HDMI} maximize the mutual information between node-level and graph-level representations. HeCo \cite{HeCo}, HeGCL\cite{HeGCL}, GTC \cite{GTC}, and MEOW \cite{MEOW} align the node representations from two tailored views. The generative HGP models \cite{HGMAE, RMR, SHAVA} aim to reconstruct the masked segments using the uncovered parts of input data. For example, HGMAE \cite{HGMAE} focuses on reconstructing both metapath-based links  and node features. RMR \cite{RMR} divides the HG into relation subgraphs and reconstructs the features of target nodes. Additionally, HGCVAE \cite{HGCVAE} combines contrastive learning and generative learning for node representation learning. To address the issue of heterophily in heterogeneous graphs (HGs), LatGRL \cite{LatGRL} introduces a similarity mining method to construct fine-grained homophilic and heterophilic latent graphs that guide representation learning.

\subsection{Self-expression}
The principle of self-expression posits that each data point can be represented as a linear combination of other data points. This property has been widely used in subspace clustering \cite{MCGC, SENet, MSESC, SAGSC}. These studies typically map the data into a latent space and exploit the self-expression property to construct an affinity matrix, which is then input into spectral clustering algorithms. Recently, some studies have explored the use of self-expression to address the problem of heterophily in graphs. In homogeneous graph learning, ROSEN \cite{ROSEN}, HEATS \cite{HEATS}, and GRAPE \cite{GRAPE} leverage the self-expressive matrix to adjust positive and/or negative samples. In heterogeneous graph learning, HERO \cite{HERO} is the most closely related work to ours. It proposes learning a self-expressive matrix to capture homophily from the subspace and neighboring nodes. However, HERO relies solely on node attributes for self-expressive learning, which limits its expressive power. In contrast, our proposed HGMS infers homophily by integrating both semantic and structural information from multiple perspectives, thereby enhancing the model's capacity to capture the complexity of relationships.

\section{PRELIMINARY}

This section introduces the basic concepts and notations of heterogeneous graphs.
\subsection{Heterogeneous Graph}

\textbf{Definition 1}.  \textbf{\textit{Heterogeneous graph}}. The heterogeneous graph (HG) can be defined as $\mathcal{G}=(\mathcal{V}, \mathcal{E}, \mathcal{A}, \mathcal{R})$, where $\mathcal{V}$ and $\mathcal{E}$ represent the sets of nodes and edges, respectively.  $\mathcal{A}$ and $\mathcal{R}$ denote sets of node types and edge types, and $|\mathcal{A}+\mathcal{R}|>2$. $\phi: \mathcal{V} \rightarrow \mathcal{A}$ is a node type mapping function.  $\varphi: \mathcal{E} \rightarrow \mathcal{R}$ is an edge type mapping function. 

\noindent \textbf{Definition 2}. \textbf{\textit{Metapath}}.  The metapath is defined as a special semantic relationship path in the form of $T_{1} \stackrel{r_{1}}{\longrightarrow} T_{2} \stackrel{r_{2}}{\longrightarrow} \ldots \stackrel{r_{l}}{\longrightarrow} T_{l+1}$ (abbreviated as $T_{1}T_{2} \cdots T_{l+1}$), where $T_{i} \in \mathcal{A}$ and $r_{i} \in $$\mathcal{R}$. 

\noindent \textbf{Definition 3}. \textbf{\textit{Metapath-based subgraph}}.  The metapath-based subgraph $G^{\mathcal{P}}=(\mathcal{V}^{\mathcal{P}},  \mathcal{E}^{\mathcal{P}})$ induced by metapath $\mathcal{P}=T_{1}T_{2} \cdots T_{l+1}$.  $G^{\mathcal{P}}$ is a homogeneous graph when $T_{1}=T_{l+1}$, which is used to capture higher-order semantics. The heterogeneous graph $\mathcal{G}=(\mathcal{V}, \mathcal{E}, \mathcal{A}, \mathcal{R})$ can be rewritten as $\mathcal{G} = (A^{\mathcal{P}_{1}}, A^{\mathcal{P}_{2}}, \cdots, A^{\mathcal{P}_{M}}; X)$, where $A^{\mathcal{P}}$ denotes the adjacency matrices of metapath-based subgraph $G^{\mathcal{P}}$, $M$ is the total number of metapath-based subgraphs, and $X$ denotes the feature matrix of the target node. 

\noindent \textbf{Definition 4}. \textbf{\textit{Metapath-based homophily ratio (MHR)}}.  Given a metapath-based subgraph $G^{\mathcal{P}}=(\mathcal{V}^{\mathcal{P}},  \mathcal{E}^{\mathcal{P}})$ and node labels $y$, we define the MHR as follows:
\begin{equation}
   \mathcal{H}\left({G}^{\mathcal{P}}\right)=\frac{1}{\left|\mathcal{E}^{\mathcal{P}}\right|} \sum_{\left(v_{i}, v_{j}\right) \in \mathcal{E}^{\mathcal{P}}} \mathds{1} \left(y_{i}=y_{j}\right),
\end{equation}
where $\mathds{1}(\cdot)$ denotes an indicator function, such that $\mathds{1}(\cdot)=1$ if the condition holds, and $\mathds{1}(\cdot)=0$ otherwise. A  metapath-based subgraph is considered homophilous when the MHR is large; otherwise, it is a heterophilic graph.

\noindent \textbf{Definition 5}. \textbf{\textit{Metapath-based connection strength (MCS)}}. 
Given a metapath $\mathcal{P}$ and two nodes $v_i$ and $v_j$, the metapath-based connection strength (MCS) represents the number of distinct paths between these two nodes along the metapath $\mathcal{P}$. MCS captures the semantic correlation between nodes. For example, given the metapath "Paper-Author-Paper" if the connection strength is 3, it indicates that at least three authors are shared between the two papers of interest.

\begin{figure}
  \centering  
  \includegraphics[scale=0.45]{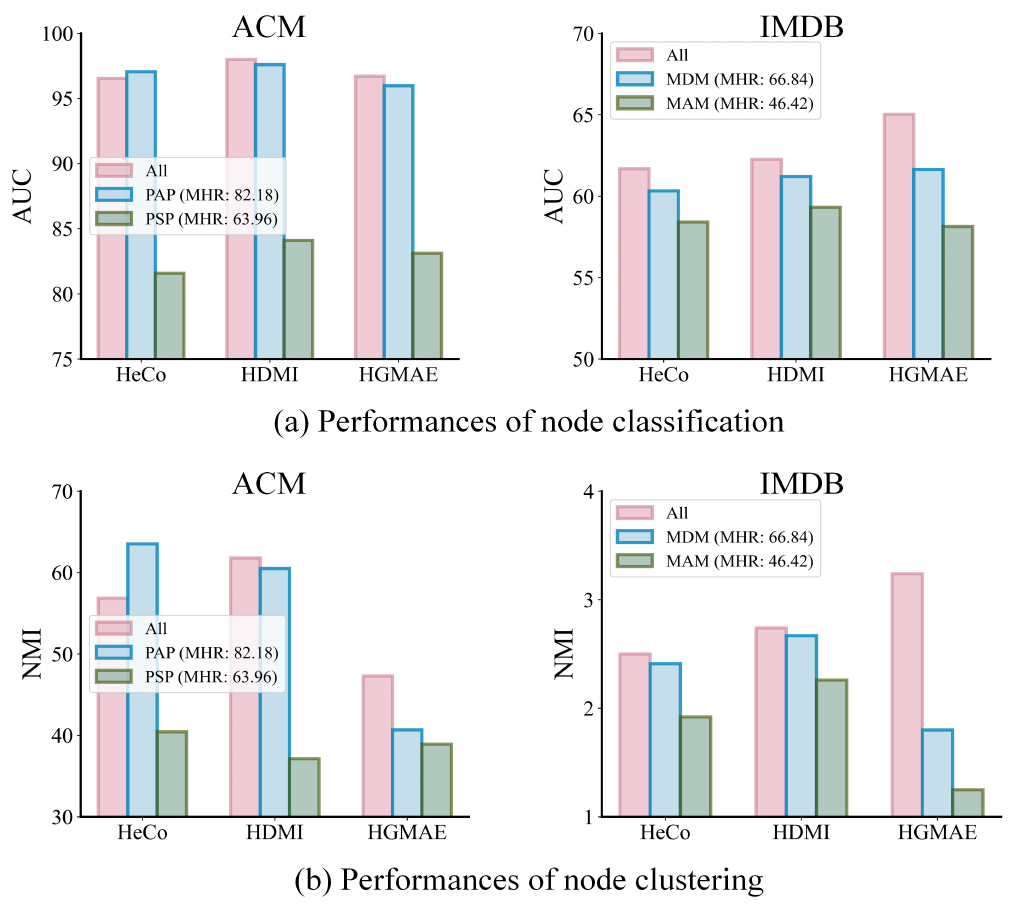}
  \caption{Impact of MHR on HGP methods.}
  \label{Expirical study 1}
\end{figure} 

\begin{figure}
  \centering  
  \includegraphics[scale=0.13]{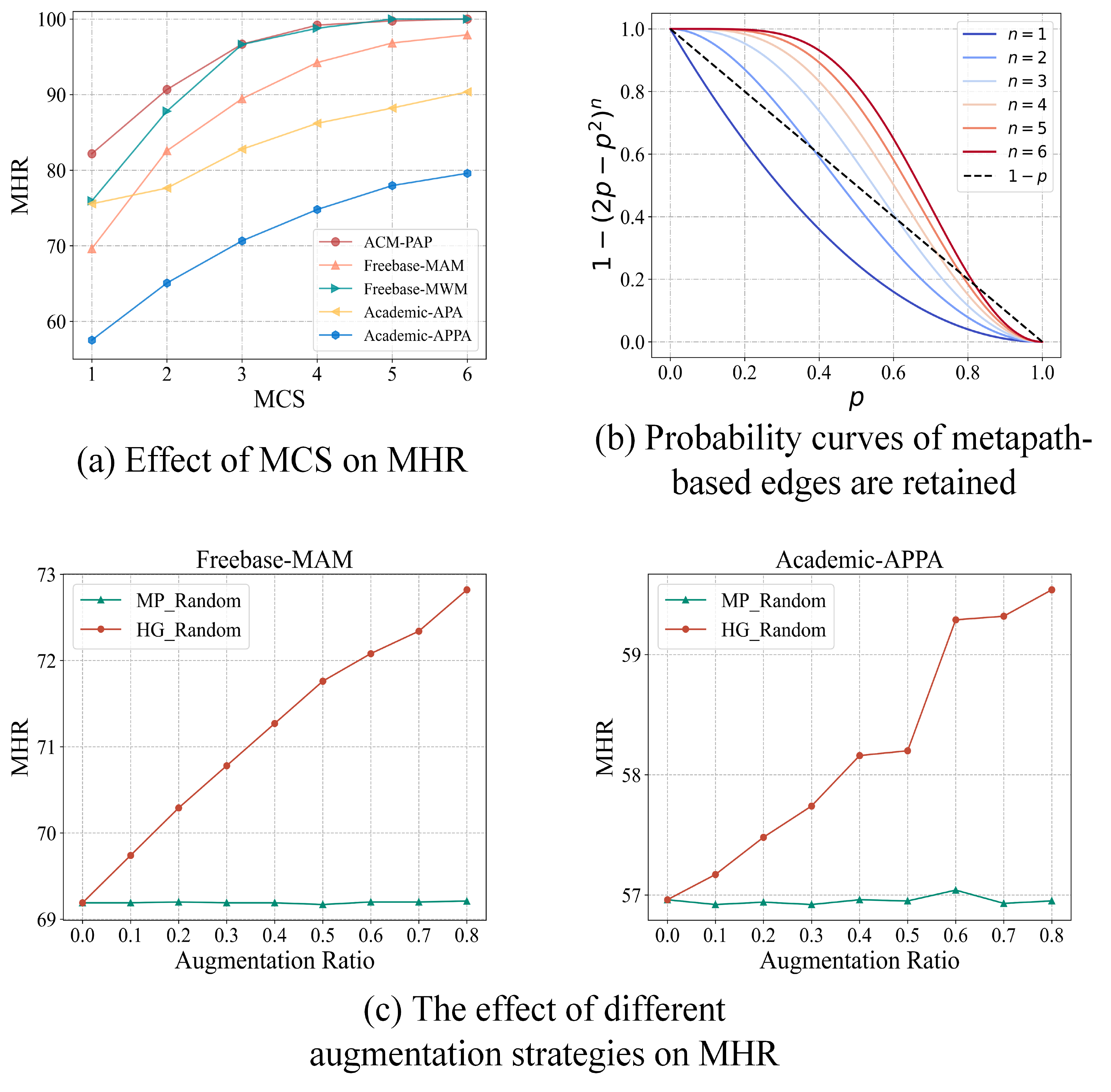}
  \caption{Studies of metapath-based connection strength. }
  \label{Expirical study 2}
\end{figure} 

\section{EMPIRICAL OBSERVATIONS}
\subsection{Empirical study 1}
\label{section: Empirical study 1}
\noindent \textbf{Experimental setup}. \textit{To investigate the relationship between metapath-based homophily ratio (MHR) and the performance of heterogeneous graph pre-training (HGP) methods,} we conduct pre-training and evaluate downstream tasks, including node classification and node clustering, on the ACM and IMDB datasets. For this empirical study, we selected two contrastive methods (i.e., HDMI and HeCo) and one generative model (i.e., HGMAE) for evaluation. Fig. \ref{Expirical study 1} illustrates the performance of the models on each metapath-based subgraph, as well as on the entire HG. 

\noindent \textbf{Observations}. Based on the results presented in Fig. \ref{Expirical study 1}, the following observations can be made: 

(1) Specifically, when using only a single metapath, higher MHR values tend to yield better performance in downstream tasks. For example, on the ACM dataset, the performance of HGP methods is significantly improved when using the PAP-based subgraph, compared to using the PSP-based subgraph. 

(2) In general, metapath-based subgraphs with lower MHR still provide valuable complementary information. The performance of HGP models can be further enhanced by integrating data from multiple subgraphs. For instance, on the IMDB dataset, HGMAE shows a significant performance boost when utilizing all available subgraphs, compared to using only a single subgraph.

Based on these findings, we conclude that \textit{it is important to enhance the homophily of metapath-based subgraphs for improving the expressive power of HGP models}.

\subsection{Empirical study 2}
\label{section: Empirical study 2}
\noindent \textbf{Experimental setup}.  To investigate the relationship between metapath-based connection strength (MCS) and metapath-based homophily ratio (MHR), we selected five metapath-based subgraphs from three datasets. The results as shown in Fig. \ref{Expirical study 2}(a). 

\noindent \textbf{Observations}. For a given metapath, its MHR increases with stronger connection strength. \textit{Therefore, leveraging structures with high MCS can enhance MHR, thereby improving the performance of HGP models.}

\section{METHODOLOGY}
Based on the observations and analysis above, we propose a novel heterogeneous graph contrastive learning framework via connection strength and multi-view self-expression, called HGMS. The framework flow diagram of HGMS is shown in Fig. \ref{Framework}.


\subsection{Augmentation Strategy of Heterogeneous Edge Dropping for Improving MHR}

In this study, we aim to design an augmentation strategy that enhances the MHR of augmented views, thereby improving the model's performance. Based on the findings of Empirical Study 2, metapath-based edges with greater connection strength tend to exhibit higher levels of homophily. When performing edge dropping, prioritizing the preservation of edges with higher connection strength and preferentially removing those with lower connection strength can effectively enhance the MHR of the augmented view. To achieve this, we propose a heterogeneous edge dropping strategy and demonstrate its feasibility below.

\begin{figure*}
  \centering  
  \includegraphics[scale=0.3]{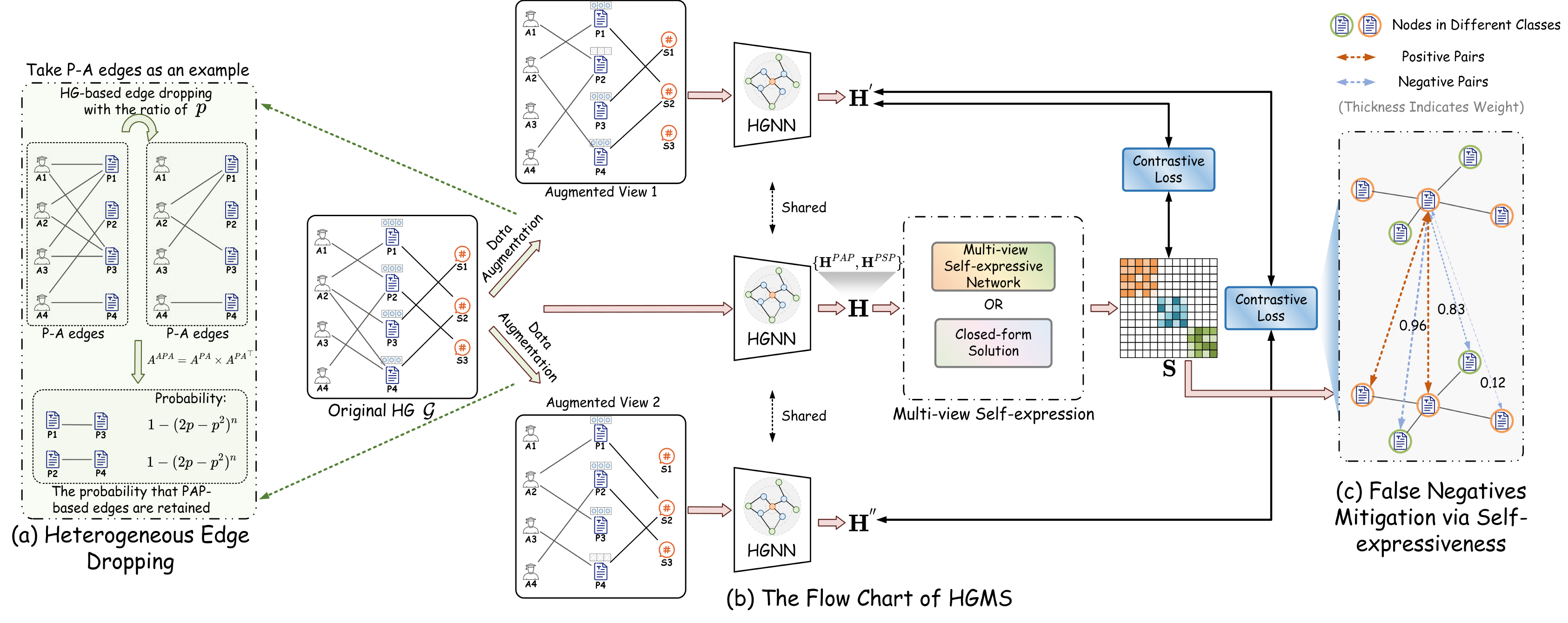}
  \caption{The overall framework of our proposed HGMS.}
  \label{Framework}
\end{figure*}

We provide a toy example in Fig. \ref{Framework}(a). In this example, we randomly drop the heterogeneous edges of the Paper-Author (P-A) type and then use the product of the corresponding adjacency matrix $A^{PA}$ to obtain the adjacency matrix of the PAP-based subgraph, i.e., $A^{APA} = A^{PA} \times {A^{PA}}^{\top}$. Given the dropping ratio of $p$ and a PAP-based edge with MCS of $n$, the probability that this edge is retained can be derived mathematically as $1-(2p-p^{2})^{n}$. The derivation process is detailed in Section \ref{section:Probability}. 

To provide an intuitive presentation, we display the probability curves in Fig. \ref{Expirical study 2}(b). When the metapath-based connection strength (MCS) $n=1$, our proposed augmentation exhibits a higher probability of these edges being discarded. As $n$ increases, the probability of retaining edges increases with higher MCS. However, if $p$ is excessively large, $1-(2p-p^{2})^{n} < 1-p$. Nevertheless, in practice, data augmentation does not usually use excessively large augmentation ratios. Overall, the heterogeneous edge dropping strategy preferentially retains edges with higher MCS, thereby improving the homophily of augmented views. 

Moreover, to further investigate the effect of graph augmentation on MHR, we selected a metapath-based subgraph from both the Freebase and Academic datasets, performed 100 perturbations on each subgraph using different augmentation strategies, and calculated the average MHR after augmentation, as shown in Fig. \ref{Expirical study 2} (c) and (d). Specifically, we considered two strategies: (1) MP$\_$Random, a metapath-based edge dropping strategy \cite{HGCMA}; and (2) HE$\_$Random, our proposed heterogeneous edge dropping. The results reveal that the MP$\_$Random has no effect on MHR, whereas HE$\_$Random consistently enhances MHR as the augmentation ratio increases.

In addition to structural augmentation, we also employ node dropping as an augmentation strategy. In Section \ref{Analysis of Data Augmentation}, we propose two additional strategies to improve the homophily of augmented views and analyze their performance.

\subsection{Heterogeneous graph encoder}

Following previous works \cite{HeCo, MEOW, HGAC}, we leverage metapath-specific GCN and semantic-level attention as the encoder for HGs. Given a HG $\mathcal{G} = (A^{\mathcal{P}_{1}}, A^{\mathcal{P}_{2}}, \cdots, A^{\mathcal{P}_{M}}; X)$, the encoding process can be formulated as:
\begin{gather}
    \hat{x}_{i} = MLP(x_{i}),  h_{i}^{\mathcal{P}_{m}}=\frac{1}{d_{i}+1} \hat{x}_{i}+\sum_{j \in \mathcal{N}_{i}^{\mathcal{P}_{m}}} \frac{1}{\sqrt{\left(d_{i}+1\right)\left(d_{j}+1\right)}} \hat{x}_{j},  \\ \label{semantic attention}
    h_{i}=\sum_{m=1}^{M} \beta_{m} \cdot h_{i}^{\mathcal{P}_{m}}, \qquad  \beta_{m} =\frac{\exp \left(\alpha_{m}\right)}{\sum_{i=1}^{M} \exp \left(\alpha_{i}\right)},\\ 
\alpha_{m} =\frac{1}{N} \sum_{i \in V} \mathbf{q}^{\top} \cdot \tanh \left(\mathbf{W} h_{i}^{\mathcal{P}_{m}}+\mathbf{b}\right),
\end{gather}
where $\mathcal{N}_{i}^{\mathcal{P}_{m}}$ denotes the neighbor set of node $i$ based on metapath $\mathcal{P}_{m}$, $d_{i}$ and $x_{i}$ are the degree and features of node $i$, respectively. $MLP(\cdot)$ represents the multilayer perceptron. $\mathbf{W} \in \mathbb{R}^{d \times d}$  and  $\mathbf{b} \in \mathbb{R}^{d \times 1}$ are learnable parameters. $\mathbf{q}$ is a semantic attention vector,  $\beta_{m}$ represents the important weight for metapath $\mathcal{P}_{m}$, $N$ denotes the number of target nodes, and $h_{i}$ denotes the final representation of the node $i$. Subsequently, we use $\mathbf{H}^{\mathcal{P}_{m}}$ denotes the node representation matrix from $\mathcal{P}_{m}$-based subgraph, and $\mathbf{H}$ denotes the final node representation matrix.

\subsection{Homophily Inference via Multi-view Self-expression}
Based on the concept that nodes in the same subspace are likely to belong to the same class \cite{subspace1}, we leverage correlations within shared subspaces to mine homophily. Specifically, we utilize the self-expression principle to model homophily within the subspace. First, we define the self-expression as follows.

\textbf{Definition 6}. \textbf{\textit{Self-expression}}. Given input data $\mathbf{X}=[\mathbf{X}_{1}, \ldots, $ $\mathbf{X}_{n}] \in \mathbb{R}^{d \times N}$ drawn from a union of an unknown number $k$ of subspaces, each data point $\mathbf{X}_{i} \in \mathbf{X}$ can be expressed as a linear combination of all other data points, i.e., $\mathbf{X}_{i}=\sum_{x_{j} \in  \mathbf{X}} s_{i j}  \mathbf{X}_{j}$, where $s_{i j}$ is the element of the self-expressive matrix $\mathbf{S}$. The matrix form can be written as: $\mathbf{X} = \mathbf{S}\mathbf{X}$. 
The self-expressive matrix $\mathbf{S}$  is learned by optimizing the following objective:
\begin{equation}
\min _{\mathbf{S}}\|\mathbf{X} - \mathbf{S}\mathbf{X}\|_{2}^{2}+\alpha \Omega({\mathbf{S}}),
\end{equation}
where $\Omega(\cdot)$ denotes the regularization function of $\mathbf{S}$, and $\alpha$ is the coefficient of the regularization term. The first term is the reconstruction loss, and the second term serves as a regularizer to avoid the trivial solution. 

Building on the subspace-preserving property \cite{HERO}, the larger the weight of $s_{i j}$, the higher the probability that points $i$ and $j$ reside within the same subspace, and thus, the greater the likelihood that they belong to the same class.

To capture semantic information from multiple perspectives, we use the learned representations $\{ \mathbf{H}^{\mathcal{P}_{m}}, 1\le m\le M\}$ derived from all metapath-based subgraphs as input for self-expressive learning. Furthermore, the optimization objective can be reformulated in a multi-view form as follows:
\begin{equation}
\min _{\mathbf{S}} \sum_{m=1}^{M} \omega^{m}\left(\|\mathbf{H}^{\mathcal{P}_{m}}-\mathbf{S}^{m}\mathbf{H}^{\mathcal{P}_{m}}\|_{2}^{2} + \alpha \Omega(\mathbf{S}^{m}) \right),
\end{equation}
where $ \omega^{m}$ is the coefficient for the $m$-th view. In this study, we use the learned attention weight $\beta_{m}$ in Eq. (\ref{semantic attention}) as the coefficient $ \omega^{m}$. Furthermore, the consensus self-expressive matrix can be obtained by fusing self-expressive matrices of all views, i.e., $\mathbf{S}=\sum_{m=1}^{M} \beta_{m} \mathbf{S}^{m}$. 

To help self-expressive matrices capture homophily, we provide attribute similarity and structural similarity as prior knowledge. Specifically, we first construct a connection strength indicator matrix $\mathbf{P} \in \mathbb{R}^{N \times N}$, where $\mathbf{P}_{i, j}=1$ if the number of metapaths (across all types) connecting node $i$ and node $j$ exceeds the threshold $\delta$, and $\mathbf{P}_{i, j}=1$ otherwise. Additionally, $\mathbf{K} \in \mathbb{R}^{N \times N}$ is a top-K graph constructed based on attribute similarity. By incorporating $\mathbf{P}$ and $\mathbf{K}$ into the regularization term, our self-expressive loss is defined as:

\begin{equation}
\label{eq: self-expression}
    \min _{\mathbf{S}} \sum_{m=1}^{M} \beta_{m}\left(\|\mathbf{H}^{\mathcal{P}_{m}}-\mathbf{S}^{m}\mathbf{H}^{\mathcal{P}_{m}}\|_{2}^{2} + \alpha \Omega(\mathbf{S}^{m}) \right) + \lambda_{1}\| \mathbf{S} -\mathbf{P} \|_{2}^{2} + \lambda_{2}\| \mathbf{S} -\mathbf{K} \|_{2}^{2},
\end{equation}
where $\lambda_{1}$ and $\lambda_{2}$ are coefficients to trade off regularization terms. 

There are several ways to solve the above objectives. Directly updating all elements $\mathbf{S}$ using the gradients is the most straightforward way, but it is computationally. In this study, we propose the HGMS framework, where the self-expressive matrix in Eq. (\ref{eq: self-expression}) is solved using two approaches: the multi-view self-expressive network, denoted as \textbf{\textit{HGMS-N}}, and the closed-form solution, denoted as \textbf{\textit{HGMS-C}}.

\textbf{Multi-view self-expressive network}. 
Inspired by \cite{SENet, MSESC}, for each node $i\in V$, we reformulate the reconstruction term as: 

\begin{equation}
 \min _{\Theta^{1}, \dots, \Theta^{M}} \sum_{m=1}^{M} \beta_{m}\|{h}_{i}^{\mathcal{P}_{m}}-\sum_{i \neq j} f_{m}\left({h}_{i}^{\mathcal{P}_{m}}, h_{j}^{\mathcal{P}_{m}} ; \Theta^{m}\right) h_{j}^{\mathcal{P}_{m}}\|_{2}^{2}, \label{self-pressive network}
\end{equation}
where $f_{m}\left({h}_{i}^{\mathcal{P}_{m}}, h_{j}^{\mathcal{P}_{m}}; \Theta^{m}\right): \mathbb{R}^{d} \times \mathbb{R}^{d} \longrightarrow \mathbb{R}$ is a learnable function parameterized by $\Theta^{m}$ for the $m$-th view. Specifically, the node representations are inputted into a two-layer MLP: ${z}_{i}^{\mathcal{P}_{m}} = \mathrm{MLP} ^{m}({h}_{i}^{\mathcal{P}_{m}})$, ${z}_{j}^{\mathcal{P}_{m}} = \mathrm{MLP} ^{m}({h}_{j}^{\mathcal{P}_{m}})$. Then, an inner product operator and a soft thresholding operator  $\Gamma_{\phi }(\cdot)$ are introduced to compute the coefficient between nodes: $f_{m}\left({h}_{i}^{\mathcal{P}_{m}}, h_{j}^{\mathcal{P}_{m}}; \Theta^{m}\right)=\Gamma_{\phi }({{z}_{i}^{\mathcal{P}_{m}}}^{\top} {{z}_{j}^{\mathcal{P}_{m}}})$, where $\mathcal{F}_{\phi}(t)=\operatorname{sgn}(t) \max (0,|t|-\phi )$, and  $\phi$ is a learnable parameter. $\operatorname{sgn}(\cdot)$ is a symbolic function that returns a positive or negative sign. The number of parameters in Eq. (\ref{self-pressive network}) is independent of the number of nodes, enabling it to scale to large graphs.

Following \cite{SENet}, we utilize the elastic net as regularization: $\Omega(\mathbf{S}^{m})=\eta |\mathbf{S}^{m}|+\frac{1-\eta}{2}(\mathbf{S}^{m})^{2}$, where $\eta \in [0, 1]$ is a balance parameter between L1 and L2 regularization. In our experiments, $\eta$ is fixed as 0.9.


\textbf{Closed-form solution}. We solve Eq. (\ref{eq: self-expression}) by setting its first-order derivative with respect to $\mathbf{S}$ equal to zero. To ensure derivability, we let $\Omega(\mathbf{S}^{m})= \sum_{i, j=1}^{N}{\mathbf{S}^{m}_{i,j}}^{2}$. The closed-form solution of $\mathbf{S}$ in Eq. (\ref{eq: self-expression}) is denoted as:
\begin{equation}
\begin{aligned}
\mathbf{S} = \left( \sum_{m=1}^{M} \beta_{m} \mathbf{H}^{\mathcal{P}_{m}} {\mathbf{H}^{\mathcal{P}_{m}}}^{\top} + \left( \alpha + \lambda_{1} + \lambda_{2} \right) \mathbf{I}_{N} \right)^{-1} \\
\left( \sum_{m=1}^{M} \beta_{m} \mathbf{H}^{\mathcal{P}_{m}} {\mathbf{H}^{\mathcal{P}_{m}}}^{\top} + \lambda_{1} \mathbf{P} + \lambda_{2} \mathbf{K} \right),
\end{aligned}
\end{equation}
where $\mathbf{I}_{N} \in \mathbb{R}^{N \times N}$ denotes the identity matrix. Based on the properties of the Block Diagonal Matrix, we concatenate multi-view representations and assign weights to them as follows: 
\begin{equation}
\mathbf{H}_{\text{cat}} = \left( \sqrt{\beta_1} \mathbf{H}^{\mathcal{P}_1} \ \ \sqrt{\beta_2} \mathbf{H}^{\mathcal{P}_2} \ \ \cdots \ \ \sqrt{\beta_M} \mathbf{H}^{\mathcal{P}_M} \right) \in \mathbb{R}^{N \times Md}.
\end{equation} 
By taking the inner product of $\mathbf{H}_{\text{cat}}$, we can obtain: $    \mathbf{H}_{\mathrm{cat}} {\mathbf{H}_{\mathrm{cat}}}^{\top}=\sum_{m=1}^{M} \beta_{m} \mathbf{H}^{\mathcal{P}_{m}} {\mathbf{H}^{\mathcal{P}_{m}}}^{\top}$. Thus, the closed-form solution for $\mathbf{S}$ simplifies to
\begin{equation}
\label{closed_form2}
\mathbf{S} = \left( \mathbf{H}_{\text{cat}} {\mathbf{H}_{\text{cat}}}^{\top} + (\alpha + \lambda_1 + \lambda_2) \mathbf{I}_{N} \right)^{-1} \left( \mathbf{H}_{\text{cat}} {\mathbf{H}_{\text{cat}}}^{\top} + \lambda_1 \mathbf{P} + \lambda_2 \mathbf{K} \right).
\end{equation} 
The time complexity of this formulation scales as the cube of the number of nodes, resulting in significant computational cost. To address this issue, we apply the Woodbury identity matrix transformation \cite{Woodbury}, which allows us to reformulate the closed-form solution of $\mathbf{S}$ as:
\begin{equation}
\begin{aligned}
\mathbf{S} = & \left( \alpha + \lambda_1 + \lambda_2 \right)^{-1} \Bigg[ \mathbf{I} - \mathbf{H}_{\text{cat}} \left( \mathbf{I}_{Md} + \mathbf{H}_{\text{cat}}^{\top} \mathbf{H}_{\text{cat}} \left( \alpha + \lambda_1 + \lambda_2 \right)^{-1} \right)^{-1} \mathbf{H}_{\text{cat}}^{\top} \Bigg] \\
& \left( \mathbf{H}_{\text{cat}} {\mathbf{H}_{\text{cat}}}^{\top} + \lambda_1 \mathbf{P} + \lambda_2 \mathbf{K} \right),
\end{aligned}
\end{equation}
where $\mathbf{I}_{Md} \in \mathbb{R}^{Md \times Md}$. This transformation reduces the time complexity of $\mathbf{S} $ to $O(N^{2}Md+M^{3}d^{3})$, where $N\gg d $. The details are further elaborated in Section \ref{section:Woodbury}. 

Finally, we perform Min-Max Normalization on each row of $\mathbf{S}$. Moreover, to ensure that the self-expressive matrix $\mathbf{S}$ remains symmetric and non-negative, we apply the following operation: $\mathbf{S}=\frac{1}{2}\left(|\mathbf{S}|+|\mathbf{S}|^{\top}\right)$.  


\subsection{False negatives mitigation via self-expression}
Although various positive sample sampling strategies for HGCL models have been proposed \cite{HeCo, HeCo++, GTC, HeGCL, HGCML}, the issue of false negatives persists. False negatives refer to nodes that belong to the same category as the anchor but are incorrectly classified as negative samples. Ideally, all false negatives could be eliminated by using labels. However, in contrastive learning, labels are not available. Since the self-expressive matrix reflects the homophily between nodes, it can be used to discriminate the authenticity of negative samples. Our tailored contrastive loss is defined as follows:
\begin{gather}
 \mathcal{L}_{s}(\mathbf{U}, \mathbf{V}) = \frac{1}{2 N} \sum_{i=1}^{N}\left(\ell_{s}\left(\boldsymbol{u}_{i}\right)+\ell_{s}\left(\boldsymbol{v}_{i}\right)\right), \label{eq:new_InfoNCE} \\
\ell_{s}\left(\boldsymbol{u}_{i}\right)=-\log \frac{\operatorname{pos}\left(\boldsymbol{u}_{i}\right)}{\operatorname{pos}\left(\boldsymbol{u}_{i}\right)+\operatorname{neg}\left(\boldsymbol{u}_{i}\right)},  \label{eq:new_InfoNCE2} \\
\operatorname{pos}\left(\boldsymbol{u}_{i}\right)=\sum_{j \in \mathbb{P}_{i}} e^{\operatorname{sim}\left(\boldsymbol{u}_{i}, \boldsymbol{v}_{j}\right) / \tau}, \\
\operatorname{neg}\left(\boldsymbol{u}_{i}\right)=\sum_{j \in \mathbb{N}_{i}} e^{\operatorname{sim}\left(\boldsymbol{u}_{i}, \boldsymbol{v}_{j}\right) \odot  (1-  \check{\mathbf{S}}_{i j})  / \tau},
\end{gather}
\begin{equation}
\check{\mathbf{S}}_{i j}=\left\{\begin{array}{ll}
1, &  \mathrm{if} \quad  \mathbf{S}_{i j} \ge  \pi_{\epsilon_{1}}(\mathbf{S}), \;  \\
\mathbf{S}_{i j},  &  \mathrm{ otherwise. }
\end{array}\right.
\end{equation}
where $\pi_{\epsilon_{1}}(\mathbf{S})$ denotes the $\epsilon_{1}$-th percentile of the values in $\mathbf{S}$. $\boldsymbol{u}_{i}$ and $\boldsymbol{v}_{i}$ denote the representations of node $i$ from the two augmented views, respectively. $\operatorname{sim}(\cdot, \cdot)$ represents the cosine similarity, and $\tau$ is a temperature coefficient. $\mathbb{P}_{i}$ and $\mathbb{N}_{i}$ represent the positive and negative sample set for node $i$, respectively. In this study, we regard metapath-based neighbors with high connection strength as positive samples, consistent with HeCo's approach \cite{HeCo}. 

Theoretically, our proposed method guarantees an improvement in the model’s performance from the perspective of mutual information (MI) maximization.

\textit{\textbf{THEOREM 1}}. \textit{The contrastive loss $ \mathcal{L}_{s}$ in Eq. (\ref{eq:new_InfoNCE}) provides a tighter lower bound of MI between input features $X$ and embeddings from two views $\mathbf{U}$ and $\mathbf{V}$, compared with the contrastive loss $ \mathcal{L}$ of previous studies \cite{HeCo, GTC}. This can be formulated as:}
\begin{equation}
    -\mathcal{L}\leqslant-\mathcal{L}_{s} \leqslant \mathcal{I}(X;U,V).
\end{equation}
The proof is provided in Section \ref{section:Theorem1}.

\subsection{Enhancement of self-expressive view}
The self-expressive matrix $\mathbf{S}$ mines homophilous latent structures. Intuitively, we can leverage the self-expressive matrix to construct an additional perspective to guide the contrastive learning process, which we refer to as the \textit{self-expressive view}. 

To sparsify $\mathbf{S}$ and filter out noise within it, we construct a purified self-expressive matrix $\hat{\mathbf{S}}$ by setting the elements in $\mathbf{S}$ that fall below a certain threshold to zero. The process is as follows:
\begin{equation}
\hat{\mathbf{S}}_{i j}=\left\{\begin{array}{ll}
0, &  \mathrm{if} \quad  \mathbf{S}_{i j} \ge  \pi_{\epsilon_{2}}(\mathbf{S}), \;  \\
\mathbf{S}_{i j},  &  \mathrm{ otherwise. }
\end{array}\right.
\end{equation}
$\pi_{\epsilon_{2}}(\mathbf{S})$ denotes the $\epsilon_{2}$-th percentile of the values in $\mathbf{S}$. Then, we learn homophilous representations via $\hat{\mathbf{S}}$, i.e., $\mathbf{H}^{S}=\hat{\mathbf{S}} \hat{X}$, where $\hat{X}$ is the node feature matrix obtained from the $MLP(\cdot)$ of the HG encoder. Furthermore, the final pre-training loss of HGMS-C is defined as:
\begin{gather}
    \mathcal{L}_{pre} = \mathcal{L}_{s}(\mathbf{H}^{'}, \mathbf{H}^{''}) +  \mathcal{L}_{s}(\mathbf{H}^{'}, \mathbf{H}^{S}),
\end{gather}
where $\mathbf{H}^{'}$ and $\mathbf{H}^{''}$ denote the node representation matrices of the two augmented views, respectively. For HGMS-N, the pre-training loss should add the self-expression loss from Eq. (\ref{eq: self-expression}), scaled by a coefficient $\mu$.

\subsection{Derivation process and proofs of theorems}

\subsubsection{The calculation of the probability that metapath-based edges are retained.}
\label{section:Probability}
Take metapath PAP in the ACM dataset as an example. Given that the discard ratio $p$ and the PAP-based connection strength between nodes $i$ and $j$ is $n$, this means there are $n$ paths of length 2 between nodes $i$ and $j$. For each path, this path fails if at least one of its two edges is discarded.  The probability that at least one edge of each path is discarded is $(2p-p^{2})$. Therefore, the probability that all paths fail is $(2p-p^{2})^{n}$. Finally, the probability that at least one path is available is $1- (2p-p^{2})^{n}$.  The derivation above fixes the metapath length to be 2. For paths of length $l\ge $ 2, we can similarly derive the probability as $1-\left(1-(1-p)^{l}\right)^{n}$.

\subsubsection{Woodbury identity matrix transformation} 
\label{section:Woodbury} 
Based on the Woodbury matrix identity \cite{Woodbury}, given four matrices, i.e., $\mathbf{A} \in \mathbb{R}^{n \times n}, \mathbf{U} \in \mathbb{R}^{n \times k}, \mathbf{C} \in \mathbb{R}^{k \times k} \text {, and } \mathbf{V} \in \mathbb{R}^{k \times n} \text {, }$ where $n$, $k$ are the dimensions of these matrices, the following matrix transformation holds: 
\begin{equation}
(\mathbf{A}+\mathbf{U C V})^{-1}=\mathbf{A}^{-1}-\mathbf{A}^{-1} \mathbf{U}\left(\mathbf{C}^{-1}+\mathbf{V A}^{-1} \mathbf{U}\right)^{-1} \mathbf{V A}^{-1}.
\end{equation} 
Without affecting generality, it is possible to replace the matrices  $\mathbf{A}$ and $\mathbf{C}$ with the identity matrix:
\begin{equation}
\label{woodbury}
(\mathbf{I}+\mathbf{U V})^{-1}=\mathbf{I}- \mathbf{U}\left(\mathbf{I}+\mathbf{V} \mathbf{U}\right)^{-1} \mathbf{V}.
\end{equation} 
According to Eq. (\ref{woodbury}), we can transform $\left( \mathbf{H}_{\text{cat}} {\mathbf{H}_{\text{cat}}}^{\top} + (\alpha + \lambda_1 + \lambda_2) \mathbf{I}_{N} \right)^{-1}$ in Eq. (\ref{closed_form2}) as:
\begin{equation}
    \left( \alpha + \lambda_1 + \lambda_2 \right)^{-1} \Bigg[ \mathbf{I} - \mathbf{H}_{\text{cat}} \left( \mathbf{I}_{Md} + \mathbf{H}_{\text{cat}}^{\top} \mathbf{H}_{\text{cat}} \left( \alpha + \lambda_1 + \lambda_2 \right)^{-1} \right)^{-1} \mathbf{H}_{\text{cat}}^{\top} \Bigg]. 
\end{equation}

\subsubsection{The proof of \textit{THEOREM} 1} 
\label{section:Theorem1}
The inequality for Theorem 1 is $  -\mathcal{L}\leqslant-\mathcal{L}_{s} \leqslant I(\mathbf{X} ; \mathbf{U}, \mathbf{V})$.

Before starting the proof, we give the contrastive loss $\mathcal{L}$  of previous studies \cite{HeCo, GTC} as follows:
\begin{gather}
  \mathcal{L}(\mathbf{U}, \mathbf{V}) = \frac{1}{2 N} \sum_{i=1}^{N}\left(\ell\left(\boldsymbol{u}_{i}\right)+\ell\left(\boldsymbol{v}_{i}\right)\right), \\
\ell\left(\boldsymbol{u}_{i}\right)=-\log \frac{\operatorname{pos}\left(\boldsymbol{u}_{i}\right)}{\operatorname{pos}\left(\boldsymbol{u}_{i}\right)+\operatorname{neg}\left(\boldsymbol{u}_{i}\right)}, \label{eq:InfoNCE1}\\
\operatorname{pos}\left(\boldsymbol{u}_{i}\right)=\sum_{j \in \mathbb{P}_{i}} e^{\operatorname{sim}\left(\boldsymbol{u}_{i}, \boldsymbol{v}_{j}\right) / \tau}, \operatorname{neg}\left(\boldsymbol{u}_{i}\right)=\sum_{j \in \mathbb{N}_{i}} e^{\operatorname{sim}\left(\boldsymbol{u}_{i}, \boldsymbol{v}_{j}\right) / \tau}, \label{eq:InfoNCE2}
\end{gather}

We first proof $  -\mathcal{L}\leqslant-\mathcal{L}_{s}$. All the elements in  $\check{\mathbf{S}}$  are between 0 and 1. Therefore, 
\begin{equation}
\sum_{j \in \mathbb{N}_{i}} e^{\operatorname{sim}\left(\boldsymbol{u}_{i}, \boldsymbol{v}_{j}\right)  / \tau} \ge \sum_{j \in \mathbb{N}_{i}} e^{\operatorname{sim}\left(\boldsymbol{u}_{i}, \boldsymbol{v}_{j}\right) \odot  (1-  \check{\mathbf{S}}_{i j})  / \tau}.
\end{equation}
In other words, the term of the negative sample $\operatorname{neg}\left(\boldsymbol{u}_{i}\right)$ of $\ell_{s}\left(\boldsymbol{u}_{i}\right)$  in Eq. (\ref{eq:new_InfoNCE2}) is less than that of $\ell\left(\boldsymbol{u}_{i}\right)$ in Eq. (\ref{eq:InfoNCE1}). And terms of the positive sample in Eq. (\ref{eq:new_InfoNCE2}) and Eq. (\ref{eq:InfoNCE1}) are consistent. Thereby, $  \ell\left(\boldsymbol{u}_{i}\right) >\ell_{s}\left(\boldsymbol{u}_{i}\right)$. Analogously, we can derive $  \ell\left(\boldsymbol{v}_{i}\right) >\ell_{s}\left(\boldsymbol{v}_{i}\right)$.  And, 
\begin{equation}
\frac{1}{2 N} \sum_{i=1}^{N}\left(\ell\left(\boldsymbol{u}_{i}\right)+\ell\left(\boldsymbol{v}_{i}\right)\right) \ge \frac{1}{2 N} \sum_{i=1}^{N}\left(\ell_{s}\left(\boldsymbol{u}_{i}\right)+\ell_{s}\left(\boldsymbol{v}_{i}\right)\right),
\end{equation}
i.e., $  -\mathcal{L}\leqslant-\mathcal{L}_{s}$. 

Next, we prove $-\mathcal{L}_{s} \leqslant \mathcal{I}(X;U,V)$. The InfoNCE \cite{infonce, infonce2} objective is formulated as: 
\begin{equation}
    I_{\mathrm{NCE}}(\mathbf{U} ; \mathbf{V}) \triangleq \mathbb{E}\left[\frac{1}{N} \sum_{i=1}^{N} \log \frac{e^{\operatorname{sim}\left(\boldsymbol{u}_{i}, \boldsymbol{v}_{i}\right)}}{\sum_{j=1}^{N} e^{\operatorname{sim}\left(\boldsymbol{u}_{i}, \boldsymbol{v}_{j}\right)}}\right],
\end{equation}
where the expectation is over $N$ samples from the joint distribution $\prod_{i} p\left(\boldsymbol{u}_{i}, \boldsymbol{v}_{i}\right)$. The proposed loss in Eq. (\ref{eq:new_InfoNCE}) is : $ \mathcal{L}_{s}(\mathbf{U}, \mathbf{V}) = \frac{1}{2} \left(\frac{1}{N} \sum_{i=1}^{N}\ell_{s}\left(\boldsymbol{u}_{i}\right)+\frac{1}{N} \sum_{i=1}^{N}\ell_{s}\left(\boldsymbol{v}_{i}\right)\right)= \frac{1}{2} \left( \mathcal{L}_{s}^{\boldsymbol{u}}(\mathbf{U}, \mathbf{V}) + \mathcal{L}_{s}^{\boldsymbol{v}}(\mathbf{U}, \mathbf{V})\right)$.  Here, the first term $\mathcal{L}_{s}^{\boldsymbol{u}}(\mathbf{U}, \mathbf{V})$ can be rewritten as: 
\begin{gather}
    \mathcal{L}_{s}^{\boldsymbol{u}}(\mathbf{U}, \mathbf{V}) =  \mathbb{E}\left[\frac{1}{N} \sum_{i=1}^{N}-\log \frac{\operatorname{pos}\left(\boldsymbol{u}_{i}\right)}{ \operatorname{pos}\left(\boldsymbol{u}_{i}\right) + \operatorname{neg}\left(\boldsymbol{u}_{i}\right)} \right],\\
   \operatorname{pos}\left(\boldsymbol{u}_{i}\right) =  \sum_{j \in \mathbb{P}_{i}} e^{\operatorname{sim}\left(\boldsymbol{u}_{i}, \boldsymbol{v}_{j}\right) / \tau}, \operatorname{neg}\left(\boldsymbol{u}_{i}\right)=\sum_{j \in \mathbb{N}_{i}} e^{\operatorname{sim}\left(\boldsymbol{u}_{i}, \boldsymbol{v}_{j}\right) \odot  (1-  \check{\mathbf{S}}_{i j})  / \tau}.
\end{gather}
As $0 \leqslant \check{\mathbf{S}}_{i j} \leqslant 1$,  $|\mathbb{P}_{i}|\geq1$ and $i \in |\mathbb{P}_{i}|$,  we have:
\begin{equation}
\begin{aligned}
    \mathcal{L}_{s}^{\boldsymbol{u}}(\mathbf{U}, \mathbf{V}) &= \mathbb{E}\left[\frac{1}{N} \sum_{i=1}^{N}-\log \frac{\operatorname{pos}\left(\boldsymbol{u}_{i}\right)}{ \operatorname{pos}\left(\boldsymbol{u}_{i}\right) + \operatorname{neg}\left(\boldsymbol{u}_{i}\right)} \right] \\
    &\geq   \mathbb{E}\left[\frac{1}{N} \sum_{i=1}^{N}-\log \frac{e^{\operatorname{sim}\left(\boldsymbol{u}_{i}, \boldsymbol{v}_{i}\right)}}{ \operatorname{pos}\left(\boldsymbol{u}_{i}\right) + \operatorname{neg}\left(\boldsymbol{u}_{i}\right)} \right]\\
      &\geq   \mathbb{E}\left[\frac{1}{N} \sum_{i=1}^{N}-\log \frac{e^{\operatorname{sim}\left(\boldsymbol{u}_{i}, \boldsymbol{v}_{i}\right)}}{ \sum_{j=1}^{N} e^{\operatorname{sim}\left(\boldsymbol{u}_{i}, \boldsymbol{v}_{j}\right)}} \right] \\
      &\geq   -I_{\mathrm{NCE}}(\mathbf{U} ; \mathbf{V}),
\end{aligned}
\end{equation}
i.e., $-\mathcal{L}_{s}^{\boldsymbol{u}}(\mathbf{U}, \mathbf{V}) \leqslant I_{\mathrm{NCE}}(\mathbf{U} ; \mathbf{V})$.  Analogously, we can also derive $-\mathcal{L}_{s}^{\boldsymbol{v}}(\mathbf{U}, \mathbf{V}) \leqslant I_{\mathrm{NCE}}(\mathbf{U} ; \mathbf{V})$. Thus, we have $\mathcal{L}_{s}(\mathbf{U}, \mathbf{V}) \leqslant  \frac{1}{2}( I_{\mathrm{NCE}}(\mathbf{U} ; \mathbf{V}) $ $+ I_{\mathrm{NCE}}(\mathbf{V} ; \mathbf{U}))$. According to \cite{infonce2}, the InfoNCE is a lower bound of MI, i.e.,  $I_{\mathrm{NCE}}(\mathbf{U} ; \mathbf{V}) \leqslant I(\mathbf{U}; \mathbf{V})$.  We then have 
\begin{equation}
    \mathcal{L}_{s}(\mathbf{U}, \mathbf{V}) \leqslant  \frac{1}{2}\left( I_{\mathrm{NCE}}(\mathbf{U} ; \mathbf{V}) + I_{\mathrm{NCE}}(\mathbf{V} ; \mathbf{U})\right) \leqslant I(\mathbf{U}; \mathbf{V}).
\end{equation}

Following \cite{Homogcl}, for  $X$, $\mathrm{U}$, $\mathbf{V}$  satisfying $I(\mathbf{U} ; \mathbf{V}) \leq I(\mathbf{U} ; \mathbf{X}) \leq I(\mathbf{X} ; \mathbf{U}, \mathbf{V})$. Thus, we finally have $ \mathcal{L}_{s} \leq I(\mathbf{X} ; \mathbf{U}, \mathbf{V})$, which concludes the proof of the second inequality.

\section{EXPERIMENTS}
\subsection{Experimental Setup}
\subsubsection{Datasets.} To verify the performance of models, we use six public benchmark datasets: ACM, DBLP, Aminer, Freebase, IMDB, and Academic. Detailed statistics of them are reported in Table \ref{tab: Datasets}.

\subsubsection{Baselines.}
We compare the proposed model with eleven strong baselines, which are categorized as follows: (1) supervised methods, including RGCN \cite{RGCN}, HAN \cite{HAN}, and HGT \cite{HGT}; (2) HG generative learning methods, including HGMAE \cite{HGMAE} and RMR \cite{RMR}; and (3) HG contrastive learning methods, including DMGI \cite{DMGI}, HDMI \cite{HDMI}, HeCo \cite{HeCo}, HeCo++ \cite{HeCo++}, HERO \cite{HERO}, and GTC \cite{GTC}.

\begin{table}[]
\caption{The statistics of the benchmark datasets.}
\label{tab: Datasets}
\renewcommand{\arraystretch}{0.5}
\tiny
\begin{tabular}{lllllll}

\toprule
Dataset  & Node  & Relations  & Metapaths     & MHR(\%)      & Target   & Classes \\ \midrule
ACM      & \begin{tabular}[c]{@{}l@{}}paper   (P): 4019\\    \\ author   (A): 7167\\    \\ subject   (S): 60\end{tabular}                            & \begin{tabular}[c]{@{}l@{}}P-A:13407\\    \\ P-S: 4019\end{tabular}                  & \begin{tabular}[c]{@{}l@{}}PAP\\    \\ PSP\end{tabular}  & \begin{tabular}[c]{@{}l@{}}80.85\\    \\ 63.93\end{tabular}          & paper  & 3\\ \midrule
DBLP      & \begin{tabular}[c]{@{}l@{}}author   (A): 4057\\    \\ paper   (P): 14328\\    \\ conference   (C): 20 \\ \\term   (T): 7723\end{tabular}                            & \begin{tabular}[c]{@{}l@{}}P-A: 19645\\    \\ P-C: 14328\\ \\ P-T: 85810 \end{tabular}                 & \begin{tabular}[c]{@{}l@{}}APA\\    \\ APCPA\\ \\ APTPA \end{tabular}  & \begin{tabular}[c]{@{}l@{}}79.86\\    \\ 66.97\\ \\ 32.45\end{tabular}          & author  & 4\\ \midrule
AMiner   & \begin{tabular}[c]{@{}l@{}}paper   (P): 6564\\    \\ author(A): 13329\\    \\ reference   (R): 35890\end{tabular}                         & \begin{tabular}[c]{@{}l@{}}P-A: 18007\\    \\ P-R: 58831\end{tabular}                 & \begin{tabular}[c]{@{}l@{}}PAP\\    \\ PRP\end{tabular}
& \begin{tabular}[c]{@{}l@{}}95.95\\    \\ 86.10\end{tabular} & paper  & 3\\ \midrule
Freebase & \begin{tabular}[c]{@{}l@{}}movie   (M): 3492\\    \\ actor   (A): 33401\\    \\ direct   (D): 2502\\    \\ writer   (W): 4459\end{tabular} & \begin{tabular}[c]{@{}l@{}}M-A: 65341\\    \\ M-D: 3762\\    \\ M-W: 6414\end{tabular} & \begin{tabular}[c]{@{}l@{}}MAM\\    \\ MDM\\    \\ MWM\end{tabular} & \begin{tabular}[c]{@{}l@{}}69.19\\    \\ 83.02\\    \\ 64.29\end{tabular} & movie  & 3 \\ \midrule
IMDB     & \begin{tabular}[c]{@{}l@{}}movie   (M): 4661\\    \\ actor   (A): 5841\\    \\ direct   (D): 2270\end{tabular}                         & \begin{tabular}[c]{@{}l@{}}M-A: 13983\\    \\ M-D: 4661\end{tabular}            & \begin{tabular}[c]{@{}l@{}}MAM\\    \\ MDM\end{tabular} & \begin{tabular}[c]{@{}l@{}}43.56\\    \\ 56.53\end{tabular}            & movie  & 3 \\ \midrule
Academic      & \begin{tabular}[c]{@{}l@{}}author   (A): 28646\\    \\ paper   (P): 21044\\    \\ venue   (V): 18 \end{tabular}                            & \begin{tabular}[c]{@{}l@{}}A-P: 69311\\    \\ P-P: 21357\\ \\ P-V: 21044 \end{tabular}                 & \begin{tabular}[c]{@{}l@{}}APA\\    \\ APPA\\ \\ APVPA \end{tabular}  & \begin{tabular}[c]{@{}l@{}}71.34\\    \\ 58.74\\ \\ 56.96\end{tabular}           & author  & 5\\ 

\bottomrule
\end{tabular}
\end{table}


\subsubsection{Implementation Detail.}
For our HGMS model, we use a 1-layer GCN to encode each metapath-based subgraph. We use grid search to tune the learning rate from 1e-3 to 1e-4, the temperature coefficient $\tau$ from 0.5 to 1.0 with intervals of 0.1, the augmentation ratio from 0.3 to 0.6 with intervals of 0.1, dropout on semantic attention from 0.0 to 0.5 with intervals of 0.05, $\epsilon_{1}$ and $\epsilon_{2}$ from 0.8 to 1.0 with intervals of 0.05, and the node dimension from \{256, 512, 1024\}. Due to space limitations, all parameter settings are provided in the open-source code repository. For all baselines, we tune their hyper-parameters based on the public code to ensure optimal performance. All results are reported as the average of 10 experimental runs conducted on a single Tesla A100 GPU with 80GB memory.

\definecolor{darkred}{rgb}{0.5, 0, 0}
\definecolor{myred}{RGB}{239,19,16}

\begin{table*}[]
\caption{Node classification performance comparison.  The highest, second highest, and third highest results are highlighted in \textcolor{myred}{bold red}, \textcolor{orange}{orange}, and \textcolor{cyan}{blue}, respectively.}
\label{tab: node classification}
\tiny
\begin{tabular}{c|c|c|ccc|cc|cccccc|cc}
\toprule
\multicolumn{3}{c|}{}                                                        & \multicolumn{3}{c|}{Supervised}     & \multicolumn{2}{c|}{Gen}            & \multicolumn{8}{c}{Contrasive}                                                    \\ \midrule
\multirow{2}{*}{Datasets} & \multirow{2}{*}{Metric} & \multirow{2}{*}{Split} & RGCN      & HAN         & HGT       & HGMAE      & RMR        & DMGI      & HDMI      & HeCo        & HeCo++    & HERO    & CTC       & \multirow{2}{*}{HGMS-N}& \multirow{2}{*}{HGMS-C}\\
                          &                         &                        & ESWC’18   & WWW’19      & WWW’20    & AAAI’22    & KDD’24     & AAAI’20   & WWW’21    & KDD’21      & TKDE’23   & ICLR’24 & NN’25     & & \\ \midrule
\multirow{9}{*}{ACM}      & \multirow{3}{*}{Ma-F1}  & 20                     & 76.21±1.9 & 85.66±2.1   & 75.06±1.7 & 88.58±0.6  & 89.02±0.3 & \textcolor{cyan}{90.78±0.3} & 89.63±0.6 & 88.56±0.8& 89.33±0.5& 88.37±1.3& 90.20±0.7 &  \textcolor{myred}{\textbf{93.11±0.4}} & \textcolor{orange}{93.07±0.4} \\
                          &                         & 40                     & 76.91±1.4 & 87.47±1.1   & 75.49±1.4 & \textcolor{cyan}{89.88±0.4}  & 88.67±0.4  & 89.92±0.4 & 89.03±0.4 & 87.61±0.5   & 88.70±0.7 &         88.59±1.1& 88.92±0.6 &  \textcolor{myred}{\textbf{92.49±0.3}} & \textcolor{orange}{92.38±0.2} \\
                          &                         & 60                     & 78.29±2.6 & 88.41±1.1   & 77.94±2.2 & 89.47±0.4  & 87.96±0.4 & \textcolor{cyan}{90.66±0.5} & 90.59±0.7 & 89.04±0.5   & 89.51±0.7 &         89.32±0.6& 89.91±0.2 & \textcolor{orange}{92.34±0.2} &  \textcolor{myred}{\textbf{92.42±0.4}} \\ \cline{2-16} 
                          & \multirow{3}{*}{Mi-F1}  & 20                     & 76.80±1.7 & 85.11±2.2   & 78.10±1.8 & 88.22±0.3  & 88.94±0.3 & 89.94±0.4 & 89.18±0.5 & 88.13±0.8   & 88.96±0.5 &         88.71±0.8& \textcolor{cyan}{90.64±0.2} & \textcolor{orange}{92.74±0.3} &  \textcolor{myred}{\textbf{92.77±0.4}} \\
                          &                         & 40                     & 79.70±1.7 & 87.21±1.2   & 77.12±1.5 & \textcolor{cyan}{89.74±0.4}  & 88.56±0.4 & 89.25±0.4 & 88.72±0.6 & 87.45±0.5   & 88.40±0.8 &         88.65±0.5& 88.55±0.3 &   \textcolor{myred}{\textbf{92.30±0.4}} & \textcolor{orange}{92.21±0.3} \\
                          &                         & 60                     & 80.40±2.4 & 88.10±1.2   & 79.60±2.5 & 89.26±0.4  & 87.76±0.4 & 89.46±0.6 & \textcolor{cyan}{90.40±0.9} & 88.71±0.5   & 89.30±0.7 &         89.62±0.4& 89.45±0.4 &  \textcolor{myred}{\textbf{92.26±0.3}} & \textcolor{orange}{92.19±0.2} \\ \cline{2-16} 
                          & \multirow{3}{*}{AUC}    & 20                     & 84.95±1.1 & 93.47±1.5   & 87.49±1.4 & 96.42±0.1  & 97.09±0.1 & 97.75±0.3 & \textcolor{cyan}{97.98±0.4} & 96.49±0.3& 97.25±0.2 &         96.48±0.4& 97.58±0.1 &  \textcolor{myred}{\textbf{98.56±0.2}} & \textcolor{orange}{98.39±0.4} \\
                          &                         & 40                     & 88.99±1.4 & 94.84±0.9   & 88.02±1.2 & 97.16±0.2  & 96.75±0.1 & 97.23±0.2 & 96.07±0.5 & 96.40±0.4& 97.08±0.2 &         96.50±0.3& \textcolor{cyan}{97.54±0.2} &  \textcolor{myred}{\textbf{98.48±0.3}} &  \textcolor{orange}{98.45±0.2} \\
                          &                         & 60                     & 91.13±1.3 & 94.68±1.4   & 90.95±1.6 & 97.21±0.1  & 95.38±0.6 & 97.72±0.4 & 97.78±0.4 & 96.55±0.3& 97.50±0.2 &         97.03±0.2& \textcolor{cyan}{97.82±0.1} & \textcolor{orange}{98.36±0.2} &  \textcolor{myred}{\textbf{98.46±0.2}} \\ \midrule
\multirow{9}{*}{DBLP}     & \multirow{3}{*}{Ma-F1}  & 20                     & 88.48±1.1          & 89.31±0.9   & 88.78±1.3          & 91.19±0.3  &            91.32±0.4& 89.94±0.4 & 90.86±0.4 & 91.28±0.2   & 91.40±0.2 &         91.68±0.3& \textcolor{cyan}{93.12±0.3} & \textcolor{orange}{93.21±0.3} &  \textcolor{myred}{\textbf{93.88±0.2}} \\
                          &                         & 40                     & 88.09±1.3          & 88.87±1.0   & 88.60±1.2          & 91.07±0.2  &            91.23±0.3& 89.25±0.4 & 90.71±0.5          & 90.34±0.3   & 90.56±0.2 &         91.33±0.3& \textcolor{orange}{91.62±0.3} &  \textcolor{myred}{\textbf{92.00±0.3}} & \textcolor{cyan}{91.21±0.2} \\
                          &                         & 60                     & 88.32±1.0          & 89.20±0.8   & 88.97±1.0          & 91.95±0.2  &            91.60±0.3& 89.46±0.6 &91.39±0.3           & 90.64±0.3   & 91.01±0.3 &         91.72±0.2& \textcolor{cyan}{92.95±0.2} & \textcolor{orange}{93.04±0.2} &  \textcolor{myred}{\textbf{93.08±0.1}} \\ \cline{2-16} 
                          & \multirow{3}{*}{Mi-F1}  & 20                     & 89.24±1.2         & 90.16±0.9   & 89.68±1.1          & 91.78±0.3  &            91.97±0.5& 90.78±0.3 &           91.44±0.5& 91.97±0.2   & 92.03±0.1 &         92.07±0.2& \textcolor{cyan}{93.67±0.3} & \textcolor{orange}{93.83±0.2} &  \textcolor{myred}{\textbf{93.94±0.3}} \\
                          &                         & 40                     & 89.02±1.0          & 89.47±0.9   & 89.32±1.2          & 91.44±0.2  &            91.15±0.3& 89.92±0.4 &           90.51±0.3& 90.76±0.3   & 90.87±0.2 &         91.14±0.3& \textcolor{orange}{92.02±0.3} &  \textcolor{myred}{\textbf{92.14±0.1}} & \textcolor{cyan}{91.81±0.1} \\
                          &                         & 60                     & 89.64±0.9          & 90.34±0.8   & 89.75±0.8          & 92.52±0.2  &            92.34±0.2& 90.66±0.5 &           91.27±0.3& 91.59±0.2   & 91.86±0.2 &         91.62±0.2& \textcolor{orange}{93.61±0.2} &  \textcolor{myred}{\textbf{93.65±0.1}} & \textcolor{cyan}{93.44±0.2} \\ \cline{2-16} 
                          & \multirow{3}{*}{AUC}    & 20                     & 97.71±0.8          & 98.07±0.6   & 97.65±0.7          & 98.54±0.1  &            98.28±0.1& 97.75±0.3 &           98.16±0.2& 98.32±0.1   & 98.39±0.1 &         98.25±0.2&  \textcolor{myred}{\textbf{98.96±0.1}} & \textcolor{cyan}{98.77±0.1} & \textcolor{orange}{98.82±0.1} \\
                          &                         & 40                     & 97.26±0.7          & 97.48±0.6   & 97.18±0.8          & 98.40±0.1  &            98.21±0.1& 97.23±0.2 &           98.01±0.1& 98.06±0.1   & 98.17±0.1 &         98.34±0.1& \textcolor{cyan}{98.46±0.1} & \textcolor{orange}{98.58±0.1} &  \textcolor{myred}{\textbf{98.70±0.1}} \\
                          &                         & 60                     & 97.59±0.6          & 97.96±0.5   & 97.68±0.5          & 98.78±0.1  &            98.56±0.1& 97.72±0.4 &           98.53±0.1& 98.59±0.1   & 98.62±0.1 &         98.44±0.1& \textcolor{cyan}{98.89±0.1} &  \textcolor{myred}{\textbf{99.12±0.1}} & \textcolor{orange}{99.01±0.1}\\ \midrule
\multirow{9}{*}{AMiner}   & \multirow{3}{*}{Ma-F1}  & 20                     & 51.73±3.5          & 56.07±3.2   & 58.90±3.1           & 72.28±0.6  &            72.41±0.7& 59.50±2.0  &           55.01±0.6& 71.38±1.1   & 72.28±1.4 &        \textcolor{cyan}{72.79±1.3}&           70.75±1.2& \textcolor{myred}{\textbf{75.16±0.9}}    & \textcolor{orange}{74.88±1.0}    \\
                          &                         & 40                     & 55.42±2.2          & 63.85±1.5   & 63.81±2.7          & 75.27±1.0  &            74.72±0.8& 61.92±2.1 &           54.10±0.9& 73.75±0.5   & 75.35±0.5 &         \textcolor{cyan}{75.62±0.8}&           72.92±0.7& \textcolor{myred}{\textbf{77.17±0.5}}    & \textcolor{orange}{76.93±0.6}    \\
                          &                         & 60                     & 55.07±1.6          & 62.02±1.2   & 64.25±2.3          & 74.67±0.6  &            75.36±0.5& 61.15±2.5 &           51.26±0.6& 75.80±1.8   & \textcolor{cyan}{76.28±0.6} &         75.17±0.6&           75.03±0.6& \textcolor{orange}{77.66±0.4}    &  \textcolor{myred}{\textbf{78.52±0.5}}    \\ \cline{2-16} 
                          & \multirow{3}{*}{Mi-F1}  & 20                     & 62.41±4.3          & 68.86±4.6   & 69.47±4.4          & 80.30±0.7  &            78.96±1.1& 63.93±3.3 &           58.71±2.2& 78.81±1.3   & 80.00±1.0 &         \textcolor{cyan}{80.68±1.3}&           78.34±1.1&  \textcolor{myred}{\textbf{82.76±1.2}}    & \textcolor{orange}{82.61±1.3}    \\
                          &                         & 40                     & 67.86±2.3           & 76.89±1.6   & 77.21±1.8          & 82.35±1.0  &            80.85±0.9& 63.60±2.5 &           63.28±2.4& 80.53±0.7   & 82.01±0.6 &         \textcolor{cyan}{82.24±0.9}&           79.76±0.8& \textcolor{orange}{84.09±0.7}    &  \textcolor{myred}{\textbf{84.12±0.9}}    \\
                          &                         & 60                     & 68.24±1.8          & 74.73±1.4   & 77.33±1.2          & 81.69±0.6  &            82.17±0.7& 62.51±2.6 &           59.16±2.3& 82.46±1.4   & \textcolor{cyan}{82.80±0.7} &         82.35±0.6&           81.85±0.5&  \textcolor{myred}{\textbf{85.18±0.4}}    & \textcolor{orange}{85.13±0.6}    \\ \cline{2-16} 
                          & \multirow{3}{*}{AUC}    & 20                     & 76.35±2.5          & 78.92±2.3   & 79.12±2.1          &   \textcolor{myred}{\textbf{93.22±0.6}}  &            90.52±0.8& 85.34±0.9 &           74.95±1.1& 90.82±0.6   & 91.59±0.6 &         90.94±0.4&           90.04±0.4& \textcolor{cyan}{92.24±0.3}    & \textcolor{orange}{92.96±0.3}    \\
                          &                         & 40                     & 78.72±2.2          & 80.72±2.1   & 80.97±1.9          &   \textcolor{myred}{\textbf{94.68±0.4}}  &            91.38±0.9& 88.02±1.3 &           74.96±1.2& 92.11±0.6   & 93.46±0.2 &         91.22±0.3&           91.97±0.5& \textcolor{orange}{93.66±0.3}    & \textcolor{cyan}{93.45±0.4}    \\
                          &                         & 60                     & 79.60±1.9          & 80.39±1.5   & 80.64±1.6          &   \textcolor{myred}{\textbf{94.59±0.3}}  &            92.14±0.4& 86.20±1.7 &           73.93±1.4& 92.40±0.7   & 93.68±0.3 &         92.14±0.2&           92.23±0.6& \textcolor{orange}{94.07±0.2}    & \textcolor{cyan}{93.79±0.3}    \\ \midrule
\multirow{9}{*}{Freebase} & \multirow{3}{*}{Ma-F1}  & 20                     &58.27±2.2           & 53.16±2.8   & 54.82±2.4          & 60.28±1.0  &            \textcolor{cyan}{60.42±1.1}& 55.79±0.9 &           55.01±0.6& 59.23±0.7   & 59.87±1.0 &         56.47±1.1& 60.40±1.5 &  \textcolor{myred}{\textbf{61.99±1.2}}    & \textcolor{orange}{61.60±1.0}          \\
                          &                         & 40                     &57.96±2.1           & 59.63±2.3   & 56.33±3.0          &  \textcolor{cyan}{62.13±0.3}&            61.51±0.7& 49.88±1.9 &           54.10±0.9& 61.19±0.6   & 61.33±0.5 &         57.29±0.8& 60.20±0.9 &  \textcolor{orange}{62.30±1.0}    &   \textcolor{myred}{\textbf{62.74±0.8}}          \\
                          &                         & 60                     &57.93±1.8           & 56.77±1.7   & 55.42±2.7          & 61.10±1.1  &            \textcolor{cyan}{61.26±0.9}& 52.10±0.7 &           51.26±0.6& 60.13±1.3   & 60.86±1.0 &         56.93±0.6& 60.81±1.2 &  \textcolor{myred}{\textbf{63.74±0.7}}    & \textcolor{orange}{62.46±0.6}          \\ \cline{2-16} 
                          & \multirow{3}{*}{Mi-F1}  & 20                     & 62.47±3.0          & 57.24±3.2   & 60.40±3.2          & 62.40±0.8&            61.27±1.0& 58.26±0.9 &           58.71±1.4& 61.72±0.6   & 62.29±1.9 &         59.73±1.8& \textcolor{cyan}{64.58±1.7} &  \textcolor{myred}{\textbf{65.96±1.4}}    & \textcolor{orange}{65.18±1.3}          \\
                          &                         & 40                     & 62.47±3.0          & 63.74±2.7   & 64.40±2.8          & 64.38±1.4&            63.79±1.2& 54.28±1.6 &           63.28±1.3& 64.03±0.7   & 64.27±0.5 &         62.10±1.6& \textcolor{cyan}{64.90±1.6} & \textcolor{orange}{65.68±1.2}    &  \textcolor{myred}{\textbf{66.30±1.0}}          \\
                          &                         & 60                     & 61.83±1.8          & 61.06±2.0   & 63.60±2.1          & 64.30±1.8  &            64.18±1.4& 56.69±1.2 &           59.16±1.3& 63.61±1.6   & 64.15±0.9 &         62.82±1.5& \textcolor{cyan}{65.86±1.3} &  \textcolor{myred}{\textbf{67.65±1.0}}    & \textcolor{orange}{66.58±0.8}          \\ \cline{2-16} 
                          & \multirow{3}{*}{AUC}    & 20                     & 74.12±2.2          & 73.26±2.1   & 70.14±2.4          & 75.03±0.7&            75.39±0.8& 73.19±1.2 &           74.95±1.3& 76.22±0.8   & \textcolor{cyan}{76.68±0.7} &         74.12±1.7& 75.21±0.9 & \textcolor{orange}{77.05±1.4}    &\textbf{  \textcolor{myred}{\textbf{77.59±0.9}}}          \\
                          &                         & 40                     & 76.53±1.4          & 77.74±1.2   & 75.15±1.6          & 74.17±0.3&            73.28±0.5& 70.77±1.6 &           74.96±1.1& 78.44±0.5   &  \textcolor{myred}{\textbf{79.51±0.3}} &         74.55±1.6& 77.10±1.5 & \textcolor{cyan}{79.14±1.1}    & \textcolor{orange}{79.30±0.8}          \\
                          &                         & 60                     & 75.38±1.2          & 75.69±1.5   & 74.07±1.4          & 78.52±0.5  &            77.84±0.6& 73.17±1.4 &           73.93±0.9& 78.04±0.4   & \textcolor{cyan}{78.27±0.7} &         74.22±1.6& 76.25±1.4 &  \textcolor{myred}{\textbf{79.24±0.8}}    & \textcolor{orange}{78.72±0.6}          \\ \midrule
\multirow{9}{*}{IMDB}     & \multirow{3}{*}{Ma-F1}  & 20                     & 43.07±2.1          & 40.32±1.3   & 40.89±2.3          & 44.32±0.8  & 44.96±0.5           & 41.66±1.1 & 44.96±1.5 & 47.76±1.6   &           48.55±1.4&         44.35±1.8&           \textcolor{cyan}{48.87±1.6}&  \textcolor{myred}{\textbf{52.21±1.2}} & \textcolor{orange}{51.86±1.1}    \\
                          &                         & 40                     & 44.57±2.5         & 42.64±1.5   & 44.23±1.8          & 46.23±1.2  & 44.77±0.4           & 44.09±1.4 & 44.77±1.2 & 46.35±1.2   &           47.46±1.3&         46.54±1.9&           \textcolor{cyan}{48.60±1.4}&  \textcolor{myred}{\textbf{55.44±0.8}}    & \textcolor{orange}{55.19±0.7}    \\
                          &                         & 60                     & 43.74±2.0          & 45.98±1.4   & 43.61±1.4          & 48.72±0.8  & 48.71±0.4           & 47.15±1.2 & 48.71±1.3 & 49.90±0.7   &           51.13±0.6&         47.29±1.1&           \textcolor{cyan}{51.28±0.8}&  \textcolor{myred}{\textbf{54.77±0.4}}    & \textcolor{orange}{54.26±0.5}    \\ \cline{2-16} 
                          & \multirow{3}{*}{Mi-F1}  & 20                     & 43.20±1.7          & 40.40±1.9   & 41.40±2.1          & 44.24±1.0  & 45.74±1.3           & 41.62±1.3 & 45.74±1.4 & 48.14±1.5   &           49.41±1.3&         45.74±1.6&           \textcolor{cyan}{49.62±1.5} &  \textcolor{myred}{\textbf{52.86±1.4}}    & \textcolor{orange}{51.40±1.2}    \\
                          &                         & 40                     &44.60±1.9           & 42.80±2.1   & 44.20±2.0          & 46.28±1.2  & 44.88±1.4           & 44.14±1.0 & 44.88±1.6 & 46.34±1.1   &           48.78±1.2&         46.44±0.6&           \textcolor{cyan}{50.33±1.2}& \textcolor{orange}{55.05±1.0}    &  \textcolor{myred}{\textbf{55.29±0.8}}    \\
                          &                         & 60                     &43.70±1.5           & 46.10±1.7   & 43.70±1.8          & 48.70±0.7  & 48.88±1.0           & 47.22±0.9 & 48.87±1.1 & 50.32±0.8   &           51.64±0.6&         48.32±0.6&           \textcolor{cyan}{51.84±0.8}&  \textcolor{myred}{\textbf{55.45±0.3}}    & \textcolor{orange}{55.01±0.3}    \\ \cline{2-16} 
                          & \multirow{3}{*}{AUC}    & 20                     &60.58±1.4           & 56.13±1.6   & 58.61±1.5          & 63.19±1.3  & 63.30±1.2           & 59.21±1.3 & 63.30±1.3 & 65.42±1.4   &           \textcolor{cyan}{66.79±1.2}&         64.126±1.2&           66.99±1.3&  \textcolor{myred}{\textbf{70.14±0.6}}    & \textcolor{orange}{69.60±0.5}    \\
                          &                         & 40                     &61.63±1.0           & 59.33±1.2   & 59.93±1.2          & 64.15±0.8  & 63.33±0.9           & 61.77±1.1 & 63.33±1.2 & 64.34±1.0   &           66.25±0.8&         64.59±1.0&           \textcolor{cyan}{67.26±0.7}&  \textcolor{myred}{\textbf{72.62±0.4}}    & \textcolor{orange}{72.52±0.3}    \\
                          &                         & 60                     &61.70±0.8           & 63.04±1.0   & 60.48±1.0          & 66.79±0.6  & 66.37±0.7           & 63.87±0.8 & 66.37±1.2 & 67.27±0.9   &           \textcolor{cyan}{68.43±1.1}&         67.46±0.8&           68.32±1.0&  \textcolor{myred}{\textbf{72.51±0.3}}    & \textcolor{orange}{72.04±0.3}    \\ \midrule
\multirow{9}{*}{Academic} & \multirow{3}{*}{Ma-F1}  & 20                     & 70.49±1.4          & 72.08 ± 1.2 & 70.81±1.2          &            68.14±1.9&            70.36±1.7& 56.41±2.6 &           68.36±1.5& 70.64 ± 1.3 & 72.09±1.4 &         70.93±1.4& \textcolor{cyan}{75.80±1.5} & \textcolor{orange}{76.33±1.1}          &            \textcolor{myred}{\textbf{76.71±0.9}}\\
                          &                         & 40                     & 72.49±1.3           & 73.24 ± 1.1 & 72.30±1.4           &            70.42±0.9&            71.69±1.3& 60.23±3.0 &           70.18±1.0& 71.52 ± 0.9 & 73.25±1.7 &         71.74±1.0& \textcolor{cyan}{77.75±1.3} & \textcolor{orange}{79.16±0.8}          &            \textcolor{myred}{\textbf{79.38±0.7}}\\
                          &                         & 60                     & 72.87±1.5          & 73.92 ± 1.7 & 72.71±1.8          &            73.55±1.2&            74.28±1.0& 61.55±2.5 &           73.62±1.1& 74.79 ± 1.2 & 75.87±1.0 &         75.44±1.1& \textcolor{cyan}{77.31±1.7} &  \textcolor{myred}{\textbf{77.94±0.7}}          &           \textcolor{orange}{77.85±0.6}\\ \cline{2-16} 
                          & \multirow{3}{*}{Mi-F1}  & 20                     & 71.82±1.3          & 73.06 ± 1.0 & 72.24±1.5          &            70.81±2.0&            72.78±1.7& 59.21±2.3 &           71.42±1.8& 72.27 ± 1.6 & 74.01±1.9 &         72.63±1.8& \textcolor{cyan}{77.03±2.1} & \textcolor{orange}{78.31±0.9}          &            \textcolor{myred}{\textbf{78.37±1.1}}\\
                          &                         & 40                     & 72.29±1.5          & 73.87 ± 1.8 & 72.58±1.6          &            72.74±1.7&            75.58±1.4& 60.22±1.6 &           73.08±1.7& 75.81 ± 1.7 & 76.50±1.5 &         74.78±1.5& \textcolor{cyan}{79.62±1.4} & \textcolor{orange}{81.05±0.7}          &            \textcolor{myred}{\textbf{81.29±0.8}}\\
                          &                         & 60                     & 72.42±1.6          & 74.01 ± 1.7 & 72.83±1.8          &            73.86±1.4&            74.49±1.5& 61.14±2.2 &           74.19±1.2& 74.81 ± 1.  & 77.15±2.2 &         74.90±1.3& \textcolor{cyan}{81.28±2.0} &  \textcolor{myred}{\textbf{81.58±0.8}}          &           \textcolor{orange}{81.43±0.8}\\ \cline{2-16} 
                          & \multirow{3}{*}{AUC}    & 20                     & 90.58±0.6          & 91.21 ± 0.3 & 90.42±0.5          &            90.68±0.8&            91.83±0.9& 83.19±1.2 &           91.33±0.6& 91.74 ± 0.4 & 92.21±0.5 &         91.69±0.7& \textcolor{cyan}{93.02±0.5} & \textcolor{orange}{95.22±0.4}          &            \textcolor{myred}{\textbf{95.45±0.3}}\\
                          &                         & 40                     & 91.22±0.7          & 92.02 ± 0.7 & 90.97±0.6          &            90.52±0.6&            91.60±0.5& 83.88±1.3 &           91.11±0.4& 91.48 ± 0.4 & 92.45±0.3 &         92.03±0.5& \textcolor{cyan}{93.43±0.6} & \textcolor{orange}{95.47±0.2}          &            \textcolor{myred}{\textbf{95.68±0.2}}\\
                          &                         & 60                     & 91.58±0.6          & 92.89 ± 0.6 & 91.32±0.9          &            90.79±0.5&            91.34±0.6& 84.21±1.4 &           91.25±0.5& 91.02 ± 0.3 & 92.56±0.7 &         92.27±0.5& \textcolor{cyan}{93.54±0.4} & \textbf{ \textcolor{myred}{95.22±0.1}}          &           \textcolor{orange}{95.13±0.2}\\ \bottomrule
\end{tabular}
\end{table*}

\begin{table}[]
\caption{Node clustering performance comparison.}
\label{tab: Node clustering}
\tiny
\begin{tabular}{c|cc|cc|cc|cc}
\toprule
Datasets & \multicolumn{2}{c|}{ACM} & \multicolumn{2}{c|}{Freebase} & \multicolumn{2}{c|}{AMiner} & \multicolumn{2}{c}{IMDB} \\ \midrule
Metrics  & NMI         & NMI        & NMI          & ARI          & NMI          & ARI          & NMI         & ARI        \\ \midrule
HGMAE    & \textcolor{cyan}{64.57}       & 67.85      & 17.18        & 17.94        & 34.70        & 34.38        & 2.76        & \textcolor{cyan}{2.62}       \\
RMR      & 54.37       & 45.65      & 18.33        & 18.15        & 30.22        & 31.59        & 2.59        & 2.47          \\ \midrule
DMGI     & 56.49       & 51.06      & 16.41        & 16.13        & 19.24        & 20.09        & 2.15        & 1.51       \\
HDMI     & 57.91       & 54.75      & 16.53        & 17.47        & 17.56        & 15.04        & 2.07        & 1.79       \\
HeCo     & 56.87       & 56.94      & 16.14        & 16.90        & 32.26        & 28.64        & 2.72        & 2.44       \\
HeCo++   & 60.82       & 60.09      & 16.67        & 17.48        & \textcolor{cyan}{38.07}        & \textcolor{cyan}{36.44}        & \textcolor{cyan}{2.92}        & 2.58           \\
HERO     & 58.41       & 57.89      & 15.34        & 15.56        & 30.11        & 31.48        & 2.19        & 2.05            \\
CTC      & 63.26       & \textcolor{cyan}{69.30}      & \textcolor{cyan}{18.70}        & \textcolor{cyan}{21.31}        & 35.21        & 34.63        & 2.86        & 2.55            \\ \midrule
HGMS-N      & \textcolor{myred}{\textbf{72.44}}       & \textcolor{myred}{\textbf{75.53}}   & \textcolor{orange}{21.35}        & \textcolor{orange}{22.49}        & \textcolor{orange}{42.87} & \textcolor{orange}{43.67} & \textcolor{myred}{\textbf{6.66}}        & \textcolor{orange}{5.25}       \\
HGMS-C      & \textcolor{orange}{71.97}       & \textcolor{orange}{74.18}   & \textcolor{myred}{\textbf{22.28}}        & \textcolor{myred}{\textbf{23.14}}        & \textcolor{myred}{\textbf{46.56}}        & \textcolor{myred}{\textbf{50.28}}        & \textcolor{orange}{6.61}        & \textcolor{myred}{\textbf{6.77}}       \\ \bottomrule
\end{tabular}
\end{table}

\subsection{Experimental Result}
\subsubsection{Node Classification.}
For the node representations obtained from our model and all baselines, we fine-tune them using a linear classifier. Following previous studies \cite{HeCo, GTC}, we select 20, 40, and 60 labeled nodes per class for training and use 1,000 nodes for both the validation and test sets. The models are evaluated using Macro-F1 (Ma-F1), Micro-F1 (Mi-F1), and AUC as performance metrics. The results are presented in Table \ref{tab: node classification}. We observe that our proposed models, HGMS-N and HGMS-C, outperform all baselines across the six datasets except for a few cases. For example, on the homophilous ACM dataset, GTC achieves Ma-F1 and Mi-F1 scores of 90.20 and 90.64 with 20 labeled nodes per class, whereas HGMS-N achieves 93.11 and 92.74, respectively. On the heterophilic IMDB dataset, HGMS-N shows improvements of 19.07\% and 19.31\% over HeCo in Ma-F1 and Mi-F1, respectively, with 40 labeled nodes per class. These results demonstrate the effectiveness of our models on both homophilous and heterophilic HGs.

\subsubsection{Node Clustering.}
We further evaluate different models through node clustering. Specifically, we apply the K-means algorithm for clustering and use common evaluation metrics, including normalized mutual information (NMI) and adjusted Rand index (ARI). The node clustering results for the four selected datasets are reported in Table \ref{tab: Node clustering}. As shown, HGMS-N and HGMS-C significantly outperform other methods, highlighting the effectiveness of HGMS. Specifically, On the homophilous ACM and Aminer datasets, HGMS-C improves NMI by 11.5\% and 22.3\%, respectively. On the heterophilic IMDB dataset, HGMS-N shows a 100\% improvement in ARI.

\subsubsection{Embedding Visualization.}
\begin{figure}
  \centering  
  \includegraphics[scale=0.12]{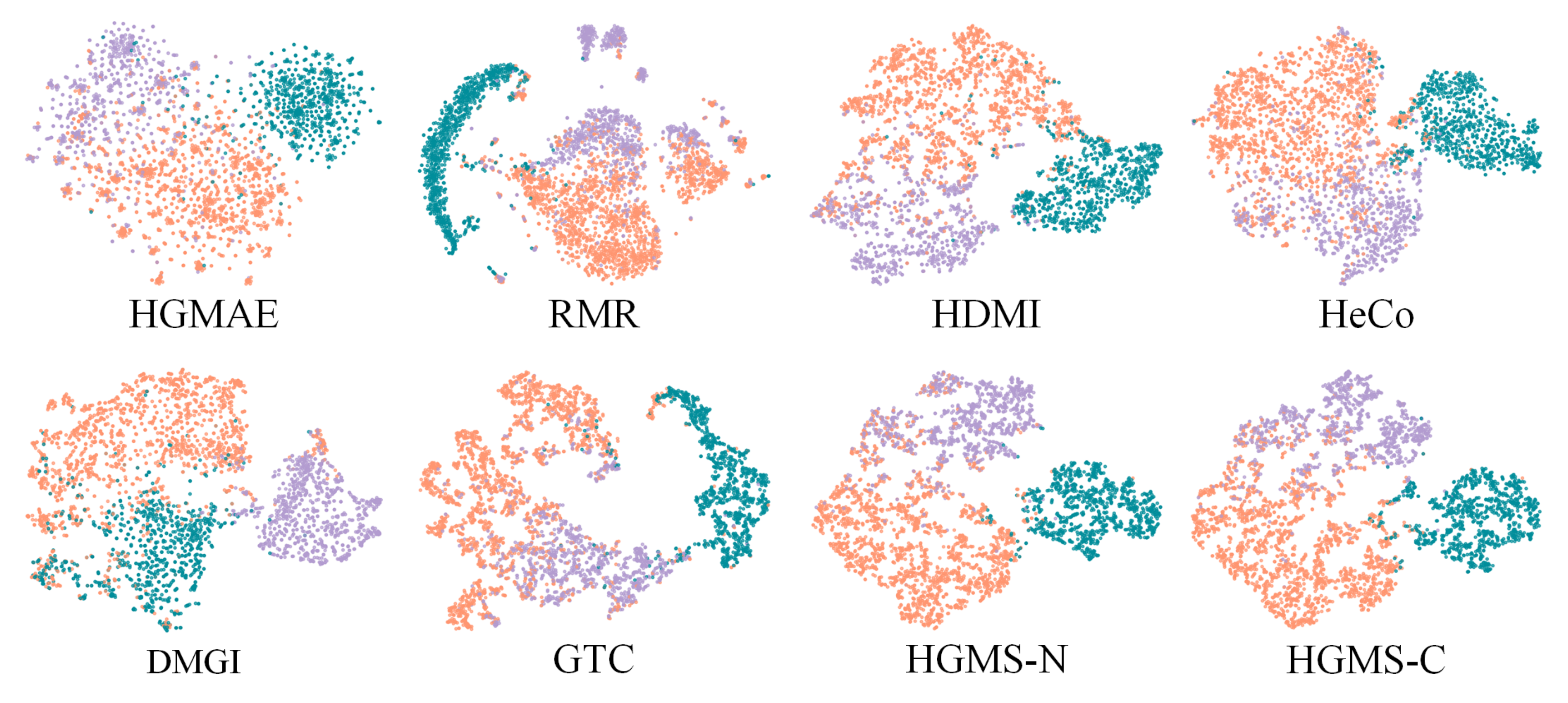}
  \caption{Visualization of the learned node embedding on ACM.}
  \label{Node Vis}
\end{figure}

For a more intuitive comparison, we visualize the distribution of node representations learned by different models using t-SNE. The visualization, in Fig. \ref{Node Vis}, uses colors to indicate the categories of the nodes. As observed, the baselines exhibit some degree of blurred boundaries between different node categories. In contrast, our HGMS-N and HGMS-C show the clearest separation, with same-class nodes forming dense clusters.

\subsubsection{Ablation Study.}
To assess the contributions of key modules, we design three variants for HGMS. \textit{w/o HGA} replaces heterogeneous edge dropping with metapath-based edge dropping. \textit{w/o SEV} removes the self-expressive view, while \textit{w/o FN} excludes the false negatives filtering module. \textit{w/o SEV \& HGA} removes both the self-expressive view and the heterogeneous edge dropping modules. We report the Ma-F1 and Mi-F1 scores with 20 labeled nodes per class, as well as NMI for the ACM, Aminer, and IMDB datasets in Table \ref{tab: ablation study}. The results show that the complete models consistently outperform the variants, demonstrating the effectiveness of the key modules. Specifically, the heterogeneous edge dropping strategy improves the homophily of augmented views. Additionally, both the self-expressive view and negative sample filtering effectively leverage the homophily extracted by the self-expressive matrix to enhance the model's expressive capacity.

\subsubsection{Impact Analysis of Data Augmentation Strategies.}
\label{Analysis of Data Augmentation}
To further explore the impact of data augmentation, we design three augmentation strategies: MP-Random, MP-PathSim, and MP-Weight. \textit{MP-Random} represents random edge dropping based on metapath. \textit{MP-PathSim} utilizes PathSim \cite{Pathsim} to compute the similarity between nodes, which is then used as the sampling probability. The sampling process follows the approach of GCA \cite{GCA}. \textit{MP-Weight} employs the metapath-based connection strength (MCS) as the sampling probability. These three strategies are applied to HGMS-N. We report the Ma-F1 and with 20 labeled nodes per class, as well as NMI for the ACM, Aminer, and IMDB datasets in Table \ref{tab: Impact of data augmentation}. The performance ranking is as follows: HGMS-N > MP-PathSim > MP-Weight > MP-Random. The results further demonstrate that retaining structures with larger MCS improves model performance, and our proposed heterogeneous edge dropping strategy outperforms the others.

\subsubsection{Visualization of Self-expressive Matrix.}
\begin{figure}
  \centering  
  \includegraphics[scale=0.1]{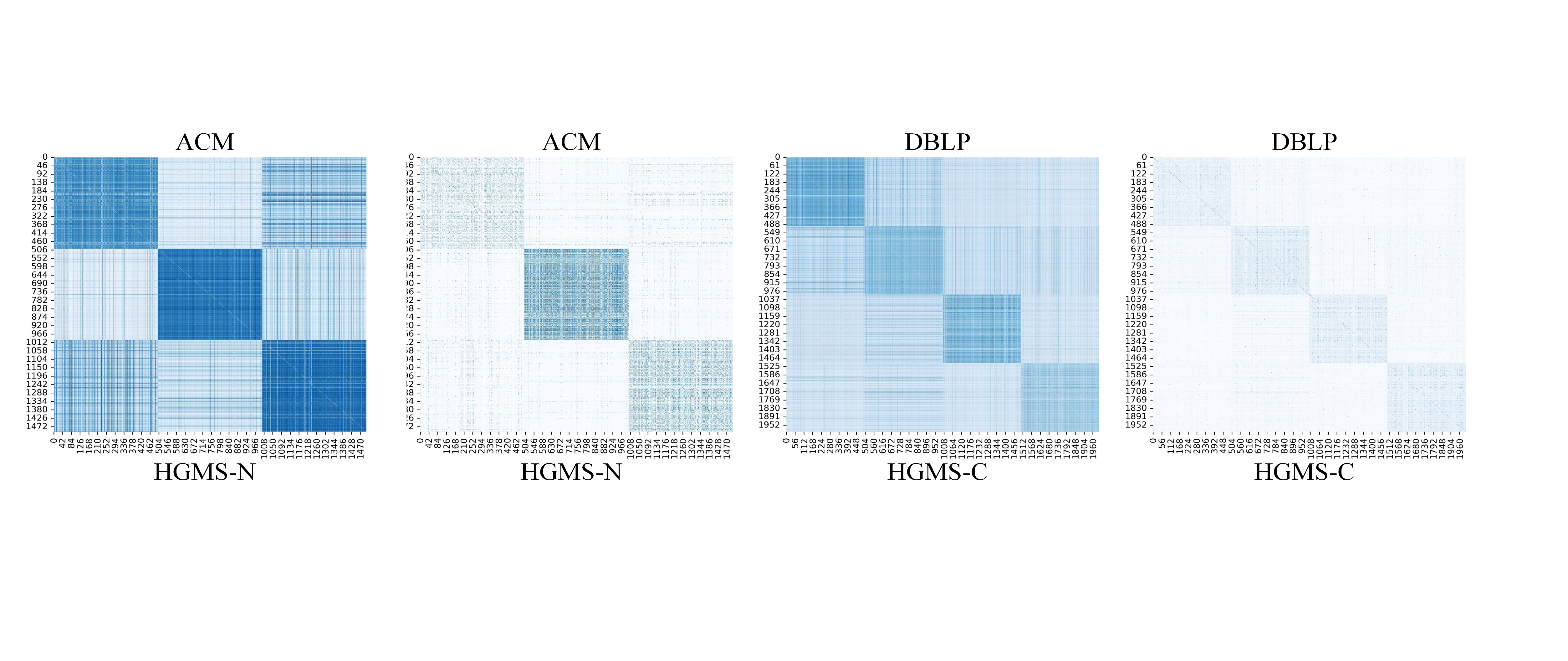}
  \caption{Visualization of the self-expressive matrices.}
  \label{self-expressive vis}
\end{figure}

To verify the effectiveness of the self-expressive matrix in inferring homophily, we visualize the heatmaps of the solved self-expressive matrices from HGMS-N and HGMS-C on the ACM and DBLP datasets. As shown in Fig. \ref{self-expressive vis}, 500 nodes are randomly selected from each category and reordered by labels. In the heatmaps, darker pixels indicate larger values in the corresponding elements of the self-expression matrix. The heatmaps reveal a block-diagonal pattern, with nodes in each block belonging to the same category. This suggests that the self-expressive matrices from both models effectively capture homophily. Furthermore, HGMS-N generates more pronounced diagonal blocks, while HGMS-C produces blocks with less noise.

\subsubsection{Robustness Analysis.}
\begin{figure}
  \centering  
  \includegraphics[scale=0.09]{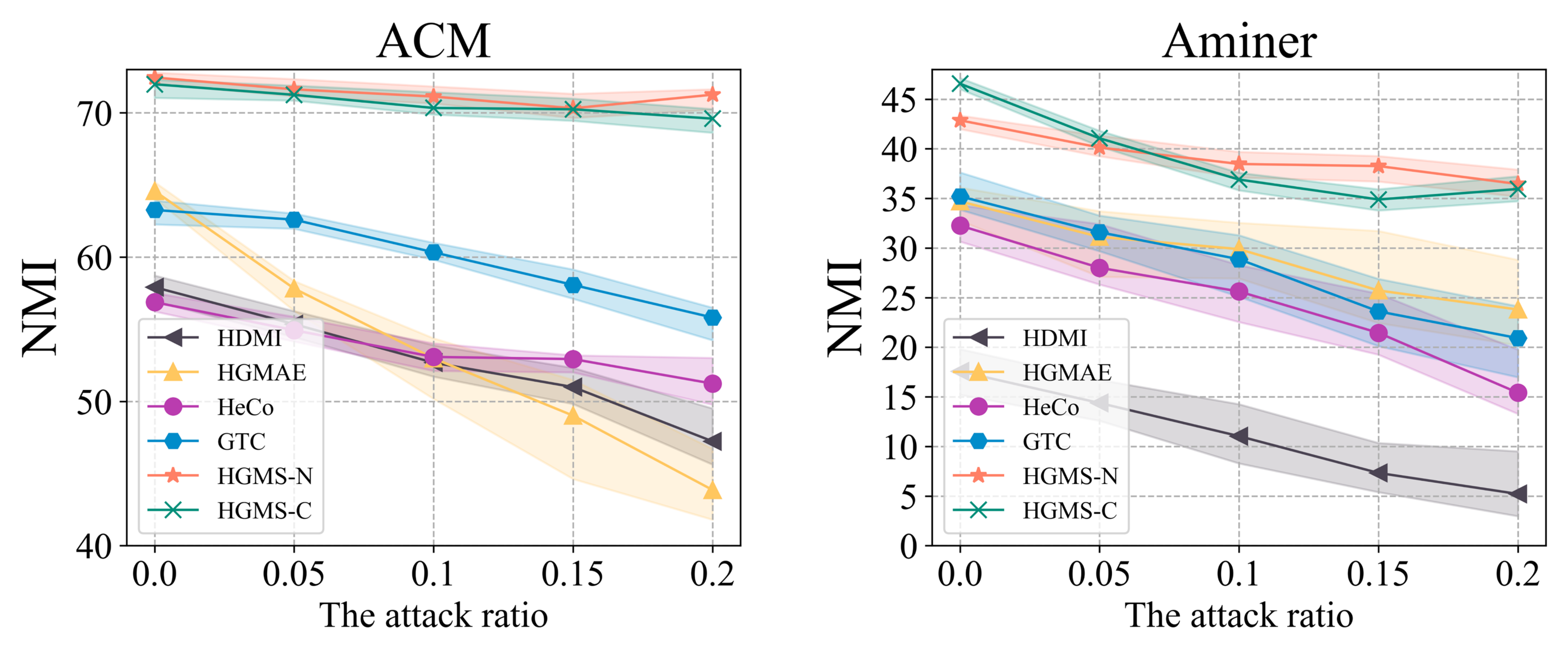}
  \caption{Performance variation of HGP models on graph data with topology attack.}
  \label{Robustness}
\end{figure}

We further evaluate the robustness of the proposed models against topology attacks. Specifically, we apply the random heterogeneous edge rewriting method \cite{HGAC} to perform a poisoning attack on heterogeneous graph pre-training models. In Fig. \ref{Robustness}, we present the node clustering performance on the ACM and IMDB datasets. The performance of the baselines degrades significantly as the attack ratio increases. In contrast, both HGMS-N and HGMS-C maintain stability in preserving their clustering results under topology attacks. This advantage can be attributed to the fact that the proposed augmentation prioritizes high-connection-strength structures, while the multi-view self-expressive learning approach consistently identifies intra-class neighbors, making the models less sensitive to structural noise.

\subsubsection{Hyper-parameter Sensitivity Analysis.}
\begin{figure}
  \centering  
  \includegraphics[scale=0.05]{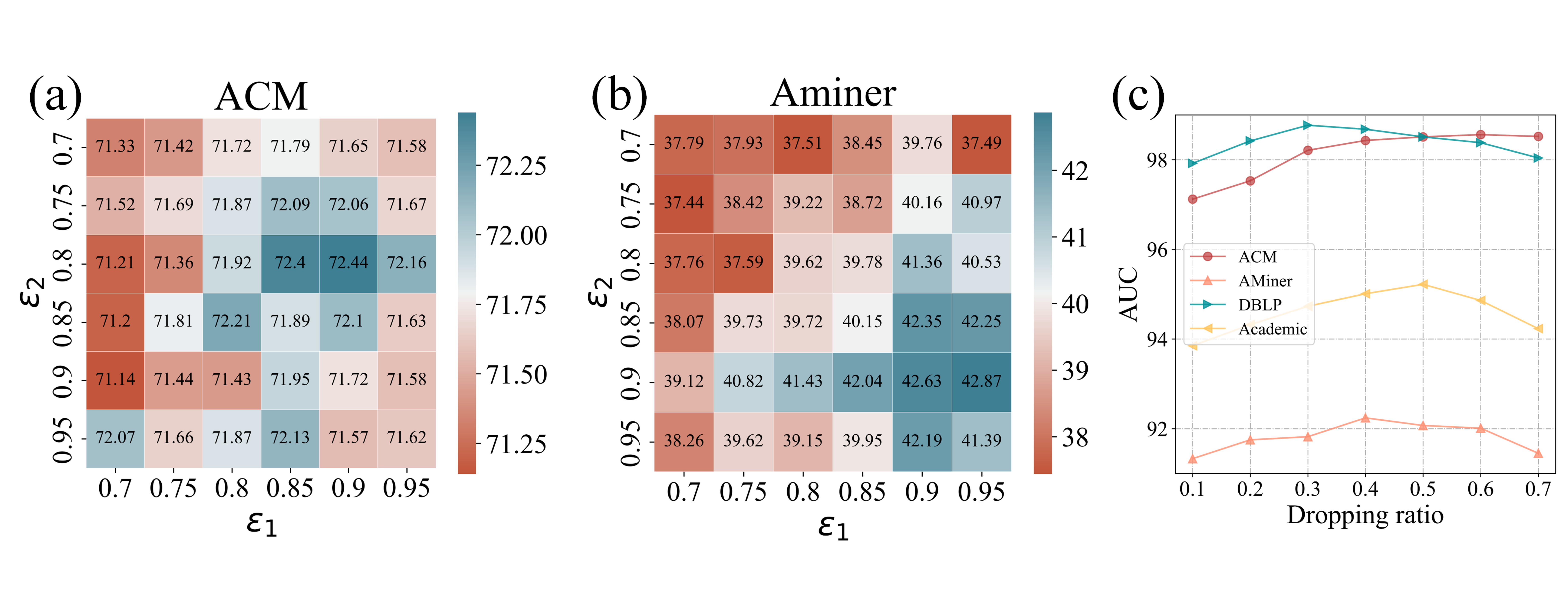}
  \caption{Hyper-parameter sensitivity analysis.}
  \label{Hyper-parameter}
\end{figure}

We investigate the sensitivity of three key hyper-parameters: the threshold values $\epsilon_{1}$ and $\epsilon_{2}$, and the augmentation ratio $p$. Specifically, $\epsilon_{1}$ and $\epsilon_{2}$ are searched within the range [0.7, 0.95] with a step size of 0.05, and the corresponding NMI scores are reported in Fig. \ref{Hyper-parameter}(a) and (b). The optimal settings are $\epsilon_{1}=0.8$ and $\epsilon_{2}=0.9$ for the ACM dataset, and $\epsilon_{1}=0.95$ and $\epsilon_{2}=0.9$ for the AMiner dataset. Overall, the performance of HGMS-N is relatively insensitive to variations in these threshold values. Furthermore, we tune the augmentation ratio $p$ from 0.1 to 0.7 with intervals of 0.1 and report the AUC score with 20 labeled nodes per class.  The optimal range of $p$ across different datasets lies between 0.3 and 0.7, consistent with the observations in Fig. \ref{Expirical study 2}(b). This range ensures the preservation of structures with high MCS in the augmented views.

\begin{table}[]
\caption{Results of ablation study.}
\label{tab: ablation study}
\tiny
\begin{tabular}{c|c|c|ccccc}
\toprule
Datasets                & Models                  & Metric & w/o HGA & w/o SEV & w/o FNF & w/o SEV\& FNF & Completed \\ \midrule
\multirow{6}{*}{ACM}    & \multirow{3}{*}{HGMS-N} & Ma-F1  & 91.71   & 92.43   & 92.85   & 92.18         & 93.11     \\
                        &                         & Mi-F1  & 92.72   & 92.07   & 92.36   & 92.03         & 92.74     \\
                        &                         & NMI    & 65.67   & 69.48   & 71.67   & 69.86         & 72.44     \\ \cline{2-8}
                        & \multirow{3}{*}{HGMS-C} & Ma-F1  & 91.59   & 92.84   & 92.82   & 92.34         & 93.07     \\
                        &                         & Mi-F1  & 92.56   & 92.31   & 92.28   & 92.19         & 92.77     \\
                        &                         & NMI    & 67.74   & 70.74   & 71.36   & 70.56         & 71.97     \\ \midrule
\multirow{6}{*}{Aminer} & \multirow{3}{*}{HGMS-N} & Ma-F1  &         73.37&         74.26&         74.52&               73.82& 75.16     \\
                        &                         & Mi-F1  &         81.62&         82.15&         81.74&               81.38& 82.76     \\
                        &                         & NMI    &         34.48&         38.24&         36.55&               37.29& 42.87\\ \cline{2-8}
                        & \multirow{3}{*}{HGMS-C} & Ma-F1  &         73.94&         73.07&         73.91&               72.62& 74.88     \\
                        &                         & Mi-F1  &         81.66&         81.56&         81.32&               80.82& 82.61     \\
                        &                         & NMI    &         45.41&         38.74&         43.36&               38.20& 46.56     \\ \midrule
\multirow{6}{*}{IMDB}   & \multirow{3}{*}{HGMS-N} & Ma-F1  &47.40         &48.35         &49.16         &48.39               & 52.21     \\
                        &                         & Mi-F1  &50.81         &46.86         &47.64         &46.69               & 52.86     \\
                        &                         & NMI    &3.35         &3.68         &3.13         &2.97               & 6.66      \\ \cline{2-8}
                        & \multirow{3}{*}{HGMS-C} & Ma-F1  &47.64    &49.03         &50.52         &49.61               & 51.86     \\
                        &                         & Mi-F1  &49.50    &49.93         &53.77         &49.81               & 51.40     \\
                        &                         & NMI    &3.02     &4.64         &5.22         &4.54               & 6.61      \\ \bottomrule
\end{tabular}
\end{table}

\begin{table}[]
\caption{Impact of data augmentation.}
\label{tab: Impact of data augmentation}
\tiny
\begin{tabular}{c|c|cccc}
\toprule
Datasets                & Metric & MP-Random & MP-PathSim & MP-Weight & HGMS-N \\ \midrule
\multirow{2}{*}{ACM}    & Ma-F1  & 91.71     & 92.14      & 91.96     & 93.11  \\
                        & NMI    & 65.67     & 67.20      & 66.24     & 72.44  \\ \midrule
\multirow{2}{*}{Aminer} & Ma-F1  & 73.37     & 74.56      & 74.52     & 75.16  \\
                        & NMI    & 34.48     & 40.67      & 39.09     & 42.87  \\ \midrule
\multirow{2}{*}{IMDB}   & Ma-F1  & 47.40     & 52.41      & 47.54     & 52.21  \\
                        & NMI    & 3.02      & 3.86       & 2.93      & 6.61   \\ \bottomrule
\end{tabular}
\end{table}

\section{CONCLUSION}
In this study, we investigate the impact of homophily in heterogeneous graphs (HGs) on the performance of heterogeneous graph pre-training (HGP) models, concluding that HGP models achieve better performance on homophilous HGs. To address this, we propose HGMS, a heterogeneous graph contrastive learning framework that leverages connection strength and multi-view self-expression to enhance the homophily of node representations. Theoretical analysis validates the effectiveness of our proposed augmentation strategy and contrastive loss. Extensive experiments across six public datasets demonstrate the superior performance of HGMS on various downstream tasks.

\bibliographystyle{ACM-Reference-Format}
\bibliography{sample-base}










\end{document}